\pdfoutput=1

\documentclass[11pt]{article}

\usepackage[preprint]{acl}

\usepackage{times}
\usepackage{latexsym}

\usepackage[T1]{fontenc}

\usepackage[utf8]{inputenc}

\usepackage{microtype}

\usepackage{inconsolata}

\usepackage{graphicx}

\usepackage{listings}
\DeclareCaptionFont{white}{\color{white}}
\DeclareCaptionFormat{listing}{%
  \parbox{\linewidth}{\colorbox{teal}{\parbox{\linewidth}{#1#2#3}}\vskip-4pt}}
\newcommand\model{\textsc{S$^2$r}}
\usepackage{xspace}
\newcommand\modelx{\textsc{S$^2$r}\xspace}
\usepackage{threeparttable}
\usepackage{amssymb}

\usepackage[utf8]{inputenc} 
\usepackage[T1]{fontenc}    
\usepackage{hyperref}       
\usepackage{url}            
\usepackage{booktabs}       
\usepackage{amsfonts}       
\usepackage{nicefrac}       
\usepackage{microtype}      
\usepackage{xcolor}         

\usepackage{boldline}
\usepackage{colortbl}
\usepackage{algorithm}
\usepackage{algorithmic}
\usepackage{color}
\usepackage{multirow}
\usepackage{amsfonts,amssymb}
\usepackage{amsmath}
\usepackage{booktabs}
\usepackage{mathtools}
\usepackage{url}
\usepackage{stfloats}
\usepackage{bbm}
\usepackage{float}
\usepackage[caption=false,font=footnotesize,farskip=0pt]{subfig}
\usepackage{amsthm}
\usepackage{enumerate}

\usepackage{enumitem}
\usepackage{makecell}
\usepackage{supertabular}
\usepackage{amsmath}
\usepackage{boldline}
\usepackage{soul}
\usepackage{natbib}
\usepackage{wrapfig}
\usepackage{xcolor}

\usepackage{graphicx}
\title{\model: Teaching LLMs to Self-verify and Self-correct \\via Reinforcement Learning
}

\author{
 \textbf{Ruotian Ma\textsuperscript{1}\thanks{~Equal contribution. This work was done during Peisong, Cheng, Jiaqi and Bang were interning at Tencent.}},
 \textbf{Peisong Wang\textsuperscript{2}\footnotemark[1]},
  \textbf{Cheng Liu\textsuperscript{1}},
 \textbf{Xingyan Liu\textsuperscript{1}},\\
 \textbf{Jiaqi Chen\textsuperscript{3}},
 \textbf{Bang Zhang\textsuperscript{1}},
  \textbf{Xin Zhou\textsuperscript{4}},
 \textbf{Nan Du\textsuperscript{1}\thanks{~Corresponding authors.}~},
 \textbf{Jia Li \textsuperscript{5}\footnotemark[2]~}
\\
\quad\quad \textsuperscript{1}Tencent\quad 
 \textsuperscript{2}Tsinghua University\\
 \textsuperscript{3}The University of Hong Kong\ \  \textsuperscript{4}Fudan University \\
 \textsuperscript{5}The Hong Kong University of Science and Technology (Guangzhou)
\\
 \small\texttt{
ruotianma@tencent.com, wps22@mails.tsinghua.edu.cn
 }
}

\begin{document}
\maketitle
\begin{abstract}
Recent studies have demonstrated the effectiveness of LLM test-time scaling. However, existing approaches to incentivize LLMs' deep thinking abilities generally require large-scale data or significant training efforts. Meanwhile, it remains unclear how to improve the thinking abilities of less powerful base models. In this work, we introduce \model, an efficient framework that enhances LLM reasoning by teaching models to self-verify and self-correct during inference. Specifically, we first initialize LLMs with iterative self-verification and self-correction behaviors through supervised fine-tuning on carefully curated data. The self-verification and self-correction skills are then further strengthened by both outcome-level and process-level reinforcement learning, with minimized resource requirements, enabling the model to adaptively refine its reasoning process during inference. 
Our results demonstrate that, with only 3.1k self-verifying and self-correcting behavior initialization samples, Qwen2.5-math-7B achieves an accuracy improvement from 51.0\% to 81.6\%, outperforming models trained on an equivalent amount of long-CoT
distilled data. 
Extensive experiments and analysis based on three base models across both in-domain and out-of-domain benchmarks validate the effectiveness of \model. Our code and data are available at \url{https://github.com/NineAbyss/S2R}.
\end{abstract}

\section{Introduction}
Recent advancements in Large Language Models (LLMs) have demonstrated a paradigm shift from scaling up training-time efforts to test-time compute \cite{snell2024scaling,kumar2024training,qi2024mutual,qwen2.5}. The effectiveness of scaling test-time compute is illustrated by OpenAI o1 \cite{o1}, which shows strong reasoning abilities by performing deep and thorough thinking, incorporating essential skills like self-checking, self-verifying, self-correcting and self-exploring during the model's reasoning process. This paradigm not only enhances reasoning in domains like mathematics and science but also offers new insights into improving the generalizability, helpfulness and safety of LLMs across various general tasks \cite{o1,guo2025deepseek}.

\begin{figure}[t]
	\centering
	\includegraphics[width=0.95\linewidth]{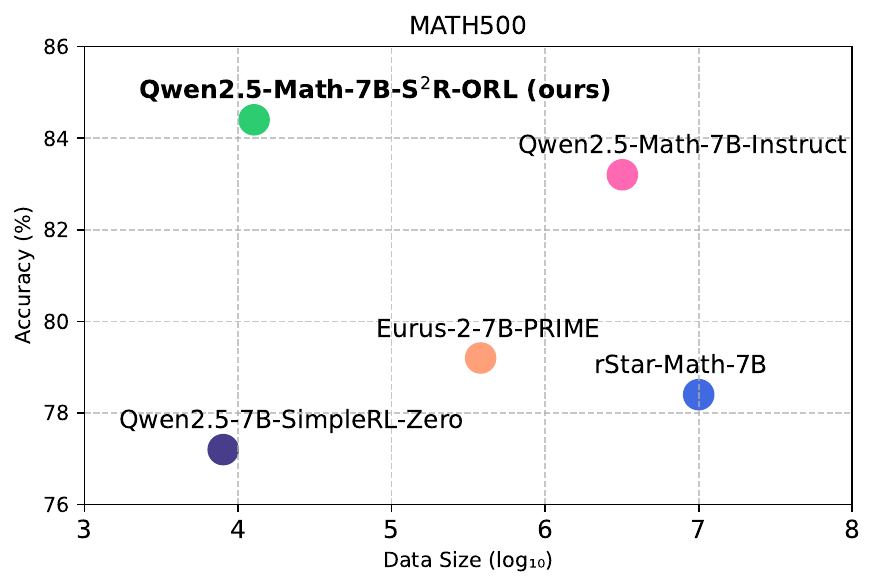}	
	\caption{The data efficiency of \modelx compared to competitive methods, with all models initialized from Qwen2.5-Math-7B.}
    \vspace{-0.3cm}
	\label{fig:datasize}
\end{figure}

Recent studies have made various attempts to replicate the success of o1. These efforts include using large-scale Monte Carlo Tree Search (MCTS) to construct long-chain-of-thought (long-CoT) training data, or to scale test-time reasoning to improve the performance 
 of current models \cite{guan2025rstar,zhao2024marco,snell2024scalingllmtesttimecompute}; constructing high-quality long-CoT data for effective behavior cloning with substantial human effort \cite{qin2024o1}; and exploring reinforcement learning to enhance LLM thinking abilities on large-scale training data and models \cite{guo2025deepseek,team2025kimi,cui2025process,yuan2024implicitprm}. Recently, DeepSeek R1 \cite{guo2025deepseek} demonstrated that large-scale reinforcement learning can incentivize LLM's deep thinking abilities, with the R1 series showcasing the promising potential of long-thought reasoning. 
However, these approaches generally requires significant resources to enhance LLMs' thinking abilities, including large datasets,
substantial training-time compute, and considerable human effort and time costs. Meanwhile, it remains unclear how to incentivize valid thinking in smaller or less powerful LLMs 
beyond distilling knowledge from more powerful models.

In this work, we propose \model, an efficient alternative to enhance the thinking abilities of LLMs, particularly for smaller or less powerful LLMs.
Instead of having LLMs imitate the thinking process of larger, more powerful models, \modelx
focus on teaching LLMs to think deeply by iteratively adopting two critical thinking skills: self-verifying and self-correcting. By acquiring these two capabilities, LLMs can continuously reassess their solutions, identify mistakes during solution exploration, and refine previous solutions after self-checking. Such a paradigm also enables flexible allocation of test-time compute to different levels of problems.
Our results show that, with only 3.1k training samples, Qwen2.5-math-7B significantly benefits from learning self-verifying and self-correcting behaviors, achieving a 51.0\% to 81.6\% accuracy improvement on the Math500 test set. 
This performance outperforms the same base model distilled from an equivalent amount of long-CoT data (accuracy 80.2\%) from QwQ-32B-Preview \cite{qwq-32b-preview}.

More importantly, \modelx employs both outcome-level and process-level reinforcement learning (RL) to further enhance the LLMs' self-verifying and self-correcting capabilities. Using only rule-based reward models, RL improves the validity of both the self-verification and self-correction process, allowing the models to perform more flexible and effective test-time scaling through a self-directed trial-and-error process. 
By comparing outcome-level and process-level RL for our task, we found that process-level supervision is particularly effective in boosting accuracy of the thinking skills at intermediate steps, which might benefit base models with limited reasoning abilities.
In contrast, outcome-level supervision enables models explore more flexible trial-and-error paths towards the correct final answer, leading to consistent improvement in the reasoning abilities of more capable base models. 
Additionally, we further show the potential of offline reinforcement learning as a more efficient alternative to the online RL training.

We conducted extensive experiments across 3 LLMs on 7 math reasoning benchmarks. Experimental results demonstrate that \modelx outperforms competitive baselines in math reasoning, including recently-released advanced o1-like models Eurus-2-7B-PRIME \cite{cui2025process}, rStar-Math-7B \cite{guan2025rstar} and Qwen2.5-7B-SimpleRL \cite{zeng2025simplerl}. We also found that \modelx is generalizable to out-of-domain general tasks, such as MMLU-PRO, highlighting the validity of the learned self-verifying and self-correcting abilities.
Additionally, we conducted a series of analytical experiments to better demonstrate the reasoning mechanisms of the obtained models, and provide insights into performing online and offline RL training for enhancing LLM reasoning.

\section{Methodology}
The main idea behind teaching LLMs self-verification and self-correction abilities is to streamline deep thinking into a critical paradigm: self-directed trial-and-error with self-verification and self-correction. 
Specifically: (1) LLMs are allowed to explore any potential (though possibly incorrect) solutions, especially when tackling difficult problems;
(2) during the process, self-verification is essential for detecting mistakes on-the-fly;
(3) self-correction enables the model to fix detected mistakes.
This paradigm forms an effective test-time scaling approach that is more accessible for less powerful base models and is generalizable across various tasks.

\begin{figure*}[t]
	\centering
	\includegraphics[width=1.0\textwidth]{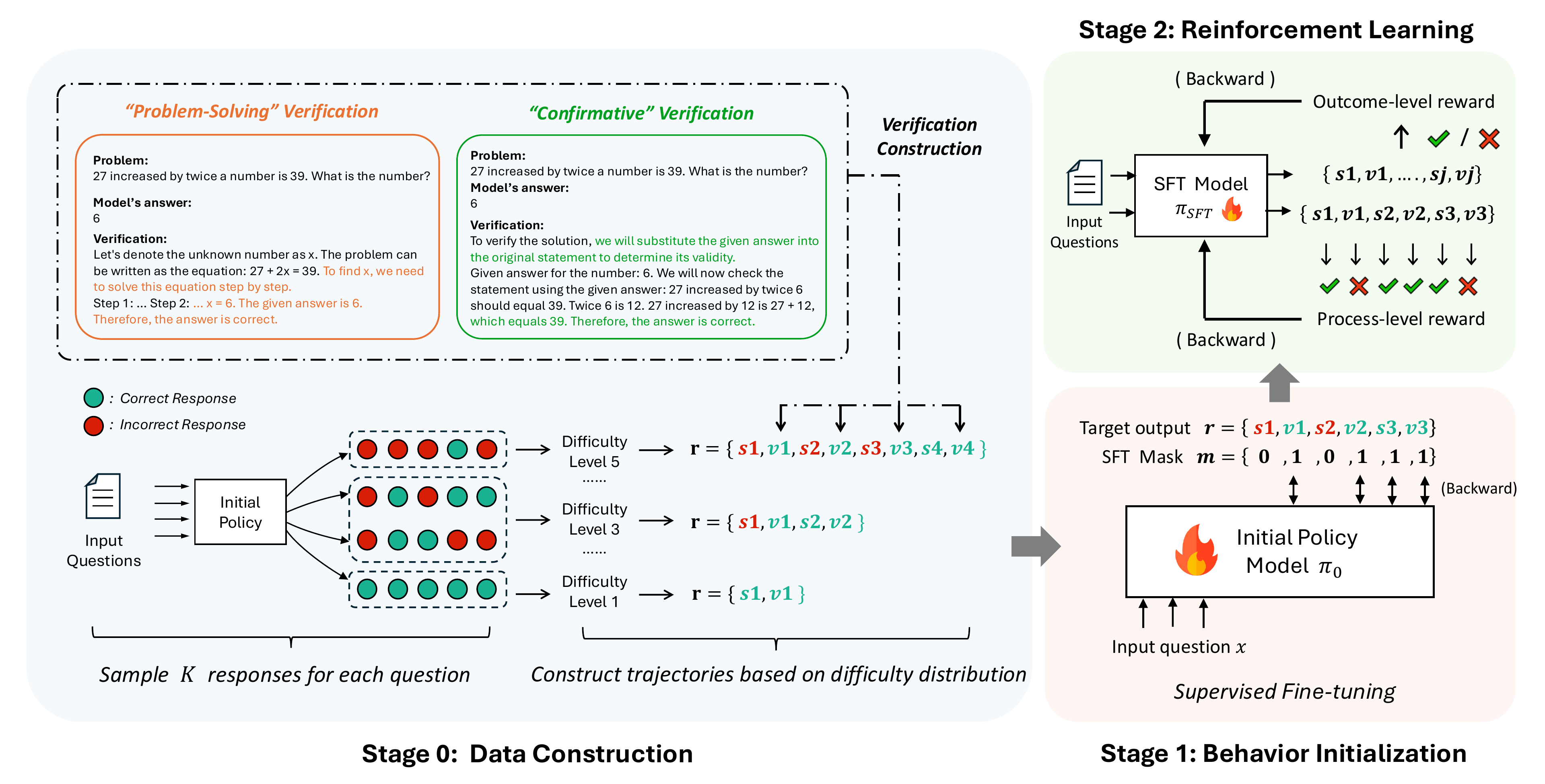}	
	\caption{Overview of \model.}
	\label{fig:method}
    \vspace{-0.3cm}
\end{figure*}

In this section, we first formally define the problem (\S \ref{sec:problem_setup}). Next, we present the two-stage training framework of \model, as described in Figure \ref{fig:method}:

\noindent \textbf{Stage 1: Behavior Initialization}: We first construct dynamic self-verifying and self-correcting trial-and-error trajectories to initialize the desired behavior. Then, we apply supervised fine-tuning (SFT) to the initial policy models using these trajectories, resulting in behavior-initialized policy models (\S \ref{sec:sft});

\noindent \textbf{Stage 2: Reinforcement Learning}: Following behavior initialization, we employ reinforcement learning to further enhance the self-verifying and self-correcting capabilities of the policy models. We explore both outcome-level and process-level RL methods, as well as their offline versions (\S \ref{sec:rl}).

\subsection{Problem Setup}\label{sec:problem_setup}

We formulate the desired LLM reasoning paradigm as a sequential decision-making process under a reinforcement learning framework.
Given a problem $x$, the language model policy $\pi$ is expected to generate a sequence of interleaved reasoning actions $y = (a_1, a_2, \cdots, a_T)$ until reaching the termination action $\texttt{<end>}$.
We represent the series of actions before an action $a_t \in y$ as $y_{:a_t}$, i.e., $y_{:a_t} = (a_1, a_2, \cdots, a_{t-i})$, where $a_t$ is excluded.
The number of tokens in $y$ is denoted as $|y|$, and the total number of actions in $y$ is denoted as $|y|_{a}$.

We restrict the action space to three types: ``\texttt{solve}'', ``\texttt{verify}'', and ``\texttt{<end>}'', where ``\texttt{solve}'' actions represent direct attempts to solve the problem, ``\texttt{verify}'' actions correspond to self-assessments of the preceding solution, and ``\texttt{<end>}'' actions signal the completion of the reasoning process. 
We denote the type of action $a_i$ as $Type(\cdot)$, where $Type(a_i) \in \{\texttt{verify}, \texttt{solve}, \texttt{<end>}\}$.
We expect the policy to learn to explore new solutions by generating ``\texttt{solve}'' actions, to self-verify the correctness of preceding solutions with ``\texttt{verify}'' actions, and to correct the detected mistakes with new ``\texttt{solve}'' actions if necessary. Therefore,
for each action $a_i$, the type of the next action $a_{i+1}$ is determined by the following rules:
\[
    \small
    Type(a_{i+1}) =
    \begin{cases}
    \texttt{verify}, & Type(a_i) = \texttt{solve} \\
    \texttt{solve}, & Type(a_i) = \texttt{verify} \\
            & \text{ and } \text{Parser}(a_i) = \textsc{incorrect} \\
    \texttt{<end>}, & Type(a_i) = \texttt{verify} \\
            & \text{ and } \text{Parser}(a_i) = \textsc{correct} \\
    \end{cases}
\]
Here, $Parser(a) \in \{\textsc{correct}, \textsc{incorrect}\}$ (for any action $a$ where $Type(a) = \texttt{verify}$ ) is a function (e.g., a regex) that converts the model's free-form verification text into binary judgments.

For simplicity, we denote the $j$-th solve action as $s_j$ and the $j$-th verify action as $v_j$.
Then we have $y = (s_1, v_1, s_2, v_2, \cdots, s_k, v_k, \texttt{<end>})$.

\subsection{Initializing Self-verification and Self-correction Behaviors}\label{sec:sft}

\subsubsection{ Learning Valid Self-verification}\label{sec:method_veri}
Learning to perform valid self-verification is the most crucial part in \model, as models can make mistakes during trial-and-error, and recognizing intermediate mistakes is critical for reaching the correct answer. In this work, we explore two methods for constructing self-verification behavior.

\paragraph{``Problem-Solving'' Verification}
The most intuitive method for verification construction is to directly query existing models to generate verifications on the policy models' responses, and then filter for valid verifications. By querying existing models using different prompts, we found that existing models tend to perform verification in a ``Problem-Solving'' manner, i.e., by re-solving the problem and checking whether the answer matches the given one. We refer to this kind of verification as ``Problem-Solving'' Verification.

\paragraph{``Confirmative'' Verification}
"Problem-solving" verification is intuitively not the ideal verification behavior we seek. Ideally, we expect the model to think outside the box and re-examine the solution from a new perspective, rather than thinking from the same problem-solving view for verification. We refer to this type of verification behavior as ``Confirmative'' Verification. Specifically, we construct ``Confirmative'' Verification by prompting existing LLMs to "verify the correctness of the answer without re-solving the problem", and filtering out invalid verifications using LLM-as-a-judge. The detail implementation can be found in Appendix \S \ref{ap:veri_implement}.

\subsubsection{Learning Self-correction}
Another critical part of \modelx is enabling the model to learn self-correction. Inspired by \citet{kumar2024training} and \citet{snell2024scalingllmtesttimecompute}, we initialize the self-correcting behavior by concatenating a series of incorrect solutions (each followed by a verification recognizing the mistakes) with a final correct solution. 
As demonstrated by \citet{kumar2024training}, LLMs typically fail to learn valid self-correction behavior through SFT, but the validity of self-correction can be enhanced through reinforcement learning. 
Therefore, we only initialize the self-correcting behavior at this stage, leaving further enhancement of the self-correcting capabilities to the RL stage. 
\subsubsection{Constructing Dynamic Trial-and-Error Trajectory}

We next construct the complete trial-and-error trajectories for behavior initialization SFT, following three principles:
\begin{itemize}[leftmargin=8pt, topsep=2pt,itemsep=2pt]
    \item To ensure the diversity of the trajectories, we construct trajectories of various lengths. Specifically, we cover $k \in \{1,2,3,4\}$ for $y=(a_1, \cdots, a_{2k})=(s_1, v_1, \cdots, s_k, v_k)$ in the trajectories.
    \item To ensure that the LLMs learn to verify and correct their own errors, we construct the failed trials in each trajectory by sampling and filtering from the LLMs' own responses.
    \item As a plausible test-time scaling method allocates reasonable effort to varying levels of problems, it is important to ensure the trial-and-error trajectories align with the difficulty level of problems. Specifically, more difficult problems will require more trial-and-error iterations before reaching the correct answer. Thus, we determine the length of each trajectory based on the accuracy of the sampled responses for each base model.
\end{itemize}

\subsubsection{Supervised Fine-tuning for Thinking Behavior Initialization}
Once the dynamic self-verifying and self-correcting training data $\mathcal{D}_{SFT}$ is ready, we optimize the policy $\pi$ for thinking behavior initialization by minimizing the following objective:
\begin{equation}
\small
\mathcal{L}=-
\mathbb{E}_{(x,y) \sim \mathcal{D}_{SFT}} \sum_{a_t \in y} \delta_{mask}(a_t) \log \pi(a_t\mid x, y_{:a_t})
\end{equation}
where the mask function $\delta_{mask}(a_t)$ for action $a_t$ in $y=(a_1, \cdots, a_T)$ is defined as:
\[
\small
\delta_{mask}(a_t)=
\begin{cases}
1, & \text{if } Type(a_t)=\texttt{verify}  \\
1, & \text{if } Type(a_t) = \texttt{solve} \text{ and } t = T-1\\
1, & \text{if } Type(a_t) = \texttt{<end>} \text{ and } t = T\\
0, & \text{otherwise}
\end{cases}
\]
That is, we optimize the probability of all verifications and only the last correct solution $s_N$ by using masks during training.

\subsection{Boosting Thinking Capabilities via Reinforcement Learning}\label{sec:rl}

After Stage 1, we initialized the policy model $\pi$ with self-verification and self-correction behavior, obtaining $\pi_{SFT}$. We then explore further enhancing these thinking capabilities of $\pi_{SFT}$ via reinforcement learning.
Specifically, we explore two simple RL algorithms: the outcome-level REINFORCE Leave-One-Out (RLOO) algorithm and a proces-level group-based RL algorithm.

\subsubsection{Outcome-level RLOO}
We first introduce the outcome-level REINFORCE Leave-One-Out (RLOO) algorithm \cite{ahmadian2024back,kool2019buy} to further enhance the self-verification and self-correction capabilities of $\pi_{SFT}$.
Given a problem $x$ and the response $y = (s_1, v_1, ..., s_T, v_T)$, we define the reward function $R_o(x, y)$ based on the correctness of the last solution $s_T$:
\[
    R_o(x, y) =
    \begin{cases}
    1,& V_{golden}(s_T) = \texttt{correct} \\
    -1, & otherwise \\
    \end{cases}
\]
Here $V_{golden}(\cdot) \in \{\texttt{correct}, \texttt{incorrect}\}$ represents ground-truth validation by matching the golden answer with the given solution.
We calculate the advantage of each response $y$ using an estimated baseline and KL reward shaping as follows:
\begin{equation}
    A(x, y) = R_o(x, y) - \hat{b} - \beta \log \frac{\pi_{\theta_{old}} (y|x)}{\pi_{ref}(y|x)}
\end{equation}
where $\beta$ is the KL divergence regularization coefficient, and $\pi_{\text{ref}}$ is the reference policy (in our case, $\pi_{SFT}$). $\hat{b}(x, y^{(m)}) = \frac{1}{M - 1} \sum_{\substack{j = 1, ..., M \\ j \neq m}}. R_o(x, y^{(j)})$ is the baseline estimation of RLOO, which represents the leave-one-out mean of $M$ sampled outputs $\{y^{(1)}, ...y^{(M)}\}$ for each input $x$, serving as a baseline estimation for each $y^{(m)}$.
Then, we optimize the policy $\pi_\theta$ by minimizing the following objective after each sampling episode based on $\pi_{\theta_{old}}$:
\begin{equation}
\begin{split}
\small
\mathcal{L}(\theta)\ & =\ - \mathbb{E}_{\substack{x \sim \mathcal{D} \\ y \sim \pi_{\theta_{\text{old}}}(\cdot | x)}}
\bigg[  \min\big( r(\theta) A(x,y), \\
        & \text{clip}\big( r(\theta), 1-\epsilon, 1+\epsilon \big) A(x,y) \big)
        \bigg]
\end{split}
\end{equation}
where $r(\theta)=\frac{\pi_{\theta}(y | x)}{\pi_{\theta_{\text{old}}}(y | x)}$ is the probability ratio.

When implementing the above loss function, we treat $y$ as a complete trajectory sampled with an input problem $x$, meaning we optimize the entire trajectory with outcome-level supervision. 
With this approach, we aim to incentivize the policy model to explore more dynamic self-verification and self-correcting trajectories on its own, which has been demonstrated as an effective practice in recent work \cite{guo2025deepseek,team2025kimi}.

\subsubsection{Process-level Group-based RL}
Process-level supervision has demonstrated effectiveness in math reasoning \cite{lightman2023let,mathshepherd}. 
 Since the trajectory of \modelx thinking is naturally divided into self-verification and self-correction processes,
it is intuitive to adopt process-level supervision for RL training.

Inspired by RLOO and process-level GRPO \cite{deepseekmath}, we designed a group-based process-level optimization method.
Specifically, we regard each action $a$ in the output trajectory $y$ as a sub-process and define the action level reward function $R_a(a \mid x, y_{:a})$ based on the action type. For each ``\texttt{solve}'' action $s_j$, we expect the policy to generate the correct solution; for each ``\texttt{verify}'' action $v_j$, we expect the verification to align with the actual solution validity. The corresponding rewards are defined as follows:
\[
\small
    R_a(s_j \mid x, y_{:s_j}) =
    \begin{cases}
    1,& V_{golden}(s_j) = \texttt{correct} \\
    -1, & otherwise \\
    \end{cases}
\]
\[
\small
    R_a(v_j \mid x, y_{:v_j}) =
    \begin{cases}
    1,& Parser(v_j) = V_{golden}(s_j) \\
    -1, & otherwise \\
    \end{cases}
\]
To calculate the advantage of each action $a_t$, we estimate the baseline as the average reward of the group of actions sharing the same \textbf{reward context}:
\[
\small
\mathbf{R}(a_t \mid x, y) = \left(R_{a}(a_i \mid x, y_{:a_i})\right)_{i=1}^{t-1}
\]
which is defined as the reward sequence of the previous actions $y_{:a_t}$ of each action $a_t$.
We denote the set of actions sharing the same reward context $\mathbf{R}(a_t \mid x, y)$ as $\mathcal{G}(\mathbf{R}(a_t \mid x, y))$.
Then the baseline can be estimated as follows:
\begin{equation}
\begin{split}
\small
    & \hat{b}(a_t \mid x, y) =  \\
    & \frac{1}{|\mathcal{G}(\mathbf{R}(a_t | x, y))|} \sum_{a \in \mathcal{G}(\mathbf{R}(a_t | x, y))} R_a(a | x^{(a)}, y^{(a)}_{:a})
\end{split}
\end{equation}
And the advantage of each action $a_t$ is:
\begin{equation}
\begin{split}
\small
    A(a_t \mid x, y) = &  R_a(a_t \mid x, y_{:a_t}) - \hat{b}(a_t \mid x, y) \\
    & - \beta \log \frac{\pi_{\theta_{old}}(a_t \mid x, y)}{\pi_{\text{ref}}(a_t \mid x, y)}
\end{split}
\end{equation}
The main idea of the group-based baseline estimation is that the actions sharing the same reward context are provided with similar amounts of information before the action is taken.
For instance, all actions sharing a reward context consisting of one failed attempt and one successful verification (i.e., $\mathbf{R}(a_t | x, y) = (-1, 1)$)
are provided with the information about the problem, a failed attempt, and the reassessment on the failure.
Given the same amount of information, it is reasonable to estimate a baseline using the average reward of these actions.

Putting it all together, we minimize the following surrogate loss function to update the policy parameters $\theta$, using trajectories collected from $\pi_{old}$:
\begin{equation}
    \small
\begin{split}
\mathcal{L}(\theta)\ & =\ - \mathbb{E}_{\substack{x \sim \mathcal{D} \\ y \sim \pi_{\theta_{\text{old}}}(\cdot | x)}}
\bigg[
\frac{1}{|y|_{a}}
\sum_{a \in y}
  \min \big( r_a(\theta) A(a| x,y_{:a}), \\
        & \text{clip}\big( r_a(\theta), 1-\epsilon, 1+\epsilon \big) A(a| x,y_{:a}) \big)
\bigg]
\end{split}
\end{equation}
where $r_a(\theta) = \frac{\pi_{\theta}(a | x, y_{:a})}{\pi_{\theta_{\text{old}}}(a | x, y_{:a})}$ is the importance ratio.

\subsection{More Efficient Training with Offline RL}\label{sec:offline_intro}

While online RL is known for its high resource requirements, offline RL, which does not require real-time sampling during training, offers a more efficient alternative for RL training. Additionally, offline sampling allows for more accurate baseline calculations with better trajectories grouping for each policy. 
As part of our exploration into more efficient RL training in \modelx framework, we also experimented with offline RL to assess its potential in further enhancing the models' thinking abilities.
In Appendix \S\ref{ap:offline_rl_details}, we include more details and formal definition for offline RL training.

\section{Experiment}
To verify the effectiveness of the proposed method, we conducted extensive experiments across 3 different base policy models on various benchmarks.

\begin{table}[t]
\centering
\renewcommand\arraystretch{1.2}
\scalebox{0.68}{
\begin{tabular}{l|cc}
\toprule
\midrule
\multicolumn{3}{l}{\textit{Stage 1: Behavior Initialization}} \\
\hline
{\bfseries Base Model} & {\bfseries Source} & {\bfseries \# Training Data} \\
\cline{1-3}
Llama-3.1-8B-Instruct &  {MATH} & {4614} \\
Qwen2-7B-Instruct &  {MATH} & {4366} \\
Qwen2.5-Math-7B &  {MATH} & {3111} \\

\midrule
\midrule
\multicolumn{3}{l}{\textit{Stage 2: Reinforcement Learning}} \\
\hline
{\bfseries Base Model} & {\bfseries Source} & {\bfseries \# Training Data} \\
\cline{1-3}
Llama-3.1-8B-Instruct &  {\ \ MATH+GSM8K}\ \  & {9601} \\
Qwen2-7B-Instruct &  {\ \ MATH+GSM8K}\ \  & {9601} \\
Qwen2.5-Math-7B &  {MATH+OpenMath2.0} & {10000} \\
\bottomrule

\end{tabular}}
\caption{Training data statistics.}
\label{tab:dataset_details}
\vspace{-0.3cm}
\end{table}

\begin{table*}[h]
	\small 
	\centering
	\resizebox{0.97\textwidth}{!}{
		\begin{tabular}%
			{@{\hskip0pt}l@{\hskip4pt}c@{\hskip6pt}c@{\hskip6pt}c@{\hskip6pt}c@{\hskip6pt}c@{\hskip6pt}c@{\hskip6pt}c@{\hskip6pt}c@{\hskip6pt}c@{\hskip0pt}}
			\toprule[1.5pt]
			&  \multicolumn{7}{c}{\bf Datasets} & \multirow{3}{*}{\makecell{\textbf{\phantom{xx}Average\phantom{xx}}}} \\
			\cmidrule{2-8} 
			\bfseries Model&\makecell{MATH\\500} &\makecell{AIME\\2024} & \makecell{AMC\\2023}& \makecell{College\\Math}& \makecell{Olympiad \\Bench}&GSM8K&\makecell{GaokaoEn\\2023}\\
			\midrule[1pt]
		  \multicolumn{9}{l}{\textit{Frontier LLMs }}\\
      GPT-4o$^{\star}$&76.6&9.3&47.5&48.5&43.3&92.9&67.5&55.1\\
		  Claude3.5-Sonnet$^{\star}$& 78.3& 16.0& -&-&-&96.4&-&- \\
		  GPT-o1-preview$^{\star}$& 85.5  &44.6&90.0&-& -& -&-&- \\
		  GPT-o1-mini$^{\star}$&90.0&56.7&95.0&57.8&65.3 &94.8&78.4&76.9 \\
		  	\midrule[1pt]
	  \multicolumn{9}{l}{\textit{Top-tier Open-source Reasoning LLMs }}\\
		  Mathstral-7B-v0.1$^{\star}$&57.8&0.0&37.5&33.7&21.5&84.9&46.0&40.2\\
		 NuminaMath-72B-CoT$^{\star}$&64.0&3.3&70.0&39.7&32.6&90.8&58.4&51.3\\
		  LLaMA3.1-70B-Instruct$^{\star}$ &65.4&23.3&50.0&42.5&27.7&94.1&54.0&51.0\\
		   Qwen2.5-Math-72B-Instruct$^{\star}$&85.6&30.0&70.0&49.5&49.0&95.9&71.9&64.6\\
		  	\midrule[1pt]
			  \multicolumn{9}{l}{\textit{General Model: Llama-3.1-8B-Instruct}}\\
		  Llama-3.1-8B-Instruct &48.0 &\underline{6.7}&\underline{30.0}&30.8 &15.6&84.4&41.0&36.6 \\
          Llama-3.1-8B-Instruct +  Original Solution SFT &31.0	&3.3	&7.5	&22.0	&8.0	&58.7&28.3&22.7\\
		  Llama-3.1-8B-Instruct + Long CoT SFT&51.4 &\underline{6.7} &27.5 &36.3  &\underline{19.0} &\underline{87.0}&\bfseries48.3&\underline{39.5} \\
         \rowcolor[rgb]{0.9, 1.0, 1.0}  \textbf{Llama-3.1-8B-\model-BI (\textit{ours})}&49.6&\bfseries10.0&20.0& 33.3&17.6&85.3&41.0&36.7 \\
              \rowcolor[rgb]{0.9, 1.0, 1.0}   \textbf{Llama-3.1-8B-\model-PRL (\textit{ours})}&\underline{53.6}&\underline{6.7}&25.0&\underline{33.7}&18.5&{86.7}&43.1& 38.2\\
	 \rowcolor[rgb]{0.9, 1.0, 1.0}   \textbf{Llama-3.1-8B-\model-ORL (\textit{ours})}&\bfseries 55.0&\underline{6.7}&\bfseries 32.5&\bfseries  34.7&\bfseries 20.7&\bfseries 87.3&\underline{45.2}&\bfseries40.3\\
		\midrule[1pt]
		\multicolumn{9}{l}{\textit{General Model: Qwen2-7B-Instruct}}\\
		Qwen2-7B-Instruct &51.2 &3.3&30.0 &18.2 &19.1&86.4&39.0&35.3 \\
        Qwen2-7B-Instruct + Original Solution SFT&  41.2	&0.0	&25.0	&30.1	&10.2	&74.5&34.8&30.8\\
        Qwen2-7B-Instruct + Long CoT SFT&60.4 &\underline{6.7} &32.5 & 36.3 &23.4 &81.2&53.5&42.0 \\
	\rowcolor[rgb]{0.9, 1.0, 1.0} 	\textbf{Qwen2-7B-\model-BI (\textit{ours})}&61.2 &3.3&27.5&\bfseries 41.1&\bfseries 27.1&\underline{ 87.4}&49.1&42.4 \\
            \rowcolor[rgb]{0.9, 1.0, 1.0}  \textbf{Qwen2-7B-\model-PRL (\textit{ours})}&\bfseries 65.4&\underline{6.7}&\underline{35.0}& \underline{36.7}&\underline{27.0}&\textbf{89.0}&\underline{49.9}&\textbf{44.2}\\
		\rowcolor[rgb]{0.9, 1.0, 1.0}  \textbf{Qwen2-7B-\model-ORL (\textit{ours})}&\underline{64.8}&3.3&\bfseries 42.5&34.7 &26.2&86.4&\bfseries 50.9&\underline{44.1}\\
		\midrule[1pt]
		\multicolumn{9}{l}{\textit{Math-Specialized Model: Qwen2.5-Math-7B}}\\
		Qwen2.5-Math-7B & 51.0 &16.7 &45.0 &21.5  &16.7&58.3&39.7&35.6 \\
		Qwen2.5-Math-7B-Instruct&83.2&13.3&72.5&47.0&40.4&\bfseries95.6&67.5&59.9 \\
        Eurus-2-7B-PRIME$^{\star}$\cite{cui2025process}&79.2&\underline{26.7}&57.8&45.0&42.1&88.0&57.1&56.6\\
        rStar-Math-7B$^{\star}$\footnotemark[2]\cite{guan2025rstar} &78.4&\underline{26.7}&47.5&\bfseries 52.5&\bfseries47.1&89.7&65.7&58.2\\
        Qwen2.5-7B-SimpleRL$^{\star}$\cite{zeng2025simplerl} &82.4&\underline{26.7}&62.5&-&43.3&-&-&-\\
        Qwen2.5-Math-7B + Original Solution SFT&58.0 &6.7 &42.5 & 35.8 &20.0 &79.5&51.9&42.1 \\
        Qwen2.5-Math-7B + Long CoT SFT&80.2 &16.7 &60.0 &\underline{49.6} &42.1 &91.4&69.1&58.4 \\
	 \rowcolor[rgb]{0.9, 1.0, 1.0}   \textbf{Qwen2.5-Math-7B-\model-BI (\textit{ours})}&81.6 &23.3 &60.0 & 43.9 &44.4&91.9&\underline{70.1}&59.3 \\
             \rowcolor[rgb]{0.9, 1.0, 1.0}  \textbf{Qwen2.5-Math-7B-\model-PRL (\textit{ours})}&\underline{83.4} &\underline{26.7}&\underline{70.0}&43.8&\underline{46.4} &\underline{93.2}&\textbf{70.4}&\underline{62.0} \\
	 \rowcolor[rgb]{0.9, 1.0, 1.0}  \textbf{Qwen2.5-Math-7B-\model-ORL (\textit{ours})}&\bfseries 84.4 &23.3&\bfseries 77.5&43.8&44.9 &92.9&\underline{70.1}&\textbf{62.4}\\
				\midrule[1pt]
\end{tabular}}

\caption{The performance of \modelx and other strong baselines on the most challenging math benchmarks is presented. \textbf{BI} refers to the behavior-initialized models through supervised fine-tuning, \textbf{ORL} denotes models trained with outcome-level RL, and \textbf{PRL} refers to models trained with process-level RL. The highest results are highlighted in \textbf{bold} and the second-best results are marked with \underline{underline}. For some baselines, we use the results from their original reports or from \citet{guan2025rstar}, denoted by $^*$.}
\label{tab:mainresults}
\vspace{-0.3cm}
\end{table*}

\subsection{Experiment Setup}

\paragraph{Base Models}

To evaluate the general applicability of our method across different LLMs, we conducted experiments using three distinct base models: Llama-3.1-8B-Instruct \cite{llama3.1}, Qwen2-7B-Instruct \cite{qwen2}, and Qwen2.5-Math-7B \cite{qwen2.5-math-7b}. Llama-3.1-8B-Instruct and Qwen2-7B-Instruct are versatile general-purpose models trained on diverse domains without a specialized focus on mathematical reasoning. In contrast, Qwen2.5-Math-7B is a state-of-the-art model specifically tailored for mathematical problem-solving and has been widely adopted in recent research on math reasoning \cite{guan2025rstar,cui2025process, zeng2025simplerl}.

\paragraph{Training Data Setup}
For {\textit{Stage 1: Behavior Initialization}}, we used the widely adopted MATH \cite{hendrycks2021measuring} training set for dynamic trial-and-error data collection \footnote[1]{We use the MATH split from \citet{lightman2023let}, i.e., 12000 problems for training and 500 problems for testing.}. For each base model, we sampled 5 responses per problem in the training data. After data filtering and sampling, we constructed a dynamic trial-and-error training set consisting of 3k-4k instances for each base model. Detailed statistics of the training set are shown in Table \ref{tab:dataset_details}. 
For \textit{Stage 2: Reinforcement Learning}, we used the MATH+GSM8K \cite{cobbe2021gsm8k} training data for RL training on the policy $\pi_{SFT}$ initialized from Llama-3.1-8B-Instruct and Qwen2-7B-Instruct.
Since Qwen2.5-math-7b already achieves high accuracy on the GSM8K training data after Stage 1, we additionally include training data randomly sampled from the OpenMath2 dataset \cite{openmath2}.
Following \cite{cui2025process}, we filter out excessively easy or difficult problems based on each $\pi_{SFT}$ from Stage 1 to enhance the efficiency and stability of RL training, resulting in RL training sets consisting of approximately 10000 instances.
Detailed statistics of the final training data can be found in Table \ref{tab:dataset_details}. Additional details on training data construction can be found in in Appendix \S \ref{ap:veri_implement}.

\paragraph{Baselines}

We benchmark our proposed method against four categories of strong baselines:

\begin{itemize}[leftmargin=8pt, topsep=2pt,itemsep=1pt]
    \item \textit{\textbf{Frontier LLMs}} includes cutting-edge proprietary models such as GPT-4o, the latest Claude, and OpenAI’s o1-preview and o1-mini. 
    \item \textit{\textbf{Top-tier open-source reasoning models}} covers state-of-the-art open-source models known for their strong reasoning capabilities, including Mathstral-7B-v0.1 \cite{mathstral}, NuminaMath-72B \cite{numina_math_datasets}, LLaMA3.1-70B-Instruct \cite{llama3.1}, and Qwen2.5-Math-72B-Instruct \cite{qwen2.5}.
    
    \item \textit{\textbf{Enhanced models built on Qwen2.5-Math-7B}}: Given the recent popularity of Qwen2.5-Math-7B as a base policy model, we evaluate \modelx against three competitive baselines that have demonstrated superior performance based on Qwen2.5-Math-7B: Eurus-2-7B-PRIME \cite{cui2025process}, rStar-Math-7B \cite{guan2025rstar}, and Qwen2.5-7B-SimpleRL \cite{zeng2025simplerl}. These models serve as direct and strong baseline for our Qwen2.5-Math-7B-based variants.
    \item \textit{\textbf{SFT with different CoT constructions}}: We also compare with training on competitive types of CoT reasoning, including the original CoT solution in the training datasets, and Long-CoT solutions distilled from QwQ-32B-Preview \cite{qwq-32b-preview}, a widely adopted open-source o1-like model \cite{chen2024not,guan2025rstar,zheng2024processbench}. Specifically, to ensure a fair comparison between behavior initialization with long-CoT and \model, we use long-CoT data of the same size as our behavior initialization data. We provide more details on the baseline data construction in Appendix \S \ref{ap:sft_base}.
\end{itemize}

More details on the baselines are included in Appendix \S \ref{ap:baseline}.

\paragraph{Evaluation Datasets}
We evaluate the proposed method on 7 diverse mathematical benchmarks. 
To ensure a comprehensive evaluation, in addition to the in-distribution GSM8K \cite{gsm8k} and MATH500 \cite{lightman2023let} test sets, we include challenging out-of-distribution benchmarks covering various difficulty levels and mathematical domains, including the AIME 2024 competition problems \cite{aime}, the AMC 2023 exam \cite{amc}, the advanced reasoning tasks from Olympiad Bench \cite{he2024olympiadbench}, and college-level problem sets from College Math \cite{mathscale}. Additionally, we assess performance on real-world standardized tests, the GaoKao (Chinese College Entrance Exam) En 2023 \cite{liao2024mario}. A detailed description of these datasets is provided in Appendix \S \ref{ap:datasets}.

\footnotetext[2] {To ensure a fair comparison, we report the Pass@1 (greedy) accuracy obtained without the process preference model of rStar, rather than the result obtained with increased test-time computation using 64 trajectories.}

\paragraph{Evaluation Metrics}
\label{sec:e_metric}
We report Pass@1 accuracy for all baselines. For inference, we employ vLLM \cite{kwon2023efficient} and develop evaluation scripts based on Qwen Math's codebase. All evaluations are performed using greedy decoding. Details of the prompts used during inference are provided in Appendix \S\ref{ap:prompts}.
All implementation details, including hyperparameter settings, can be found in Appendix \S\ref{ap:hyper}.
\vspace{-0.1mm}

\subsection{Main Results}

Table \ref{tab:mainresults} shows the main results of \modelx compared with baseline methods. We can observe that:
(1) \modelx consistently improves the reasoning abilities of models across all base models. Notably, on Qwen2.5-Math-7B, the proposed method improves the base model by 32.2\% on MATH500 and by 34.3\% on GSM8K.
(2) Generally, \modelx outperforms the baseline methods derived from the same base models across most benchmarks. 
Specifically, on Qwen2.5-Math-7B, \modelx surpasses several recently proposed competitive baselines, such as Eurus-2-7B-PRIME, rStar-Math-7B and Qwen2.5-7B-SimpleRL. While Eurus-2-7B-PRIME and rStar-Math-7B rely on larger training datasets (Figure \ref{fig:datasize}) and require more data construction and reward modeling efforts, \modelx only needs linear sampling efforts for data construction, 10k RL training data and rule-based reward modeling. 
These results highlight the efficiency of \model.
(3) With the same scale of SFT data, \modelx also outperforms the long-CoT models distilled from QwQ-32B-Preview, demonstrating that learning to self-verify and self-correct is an effective alternative to long-CoT for test-time scaling in smaller LLMs.

\noindent\textbf{Comparing process-level and outcome-level RL},  we find that outcome-level RL generally outperforms process-level RL across the three models. This is likely because outcome-level RL allows models to explore trajectories without emphasizing intermediate accuracy, which may benefit enhancing long-thought reasoning in stronger base models like Qwen2.5-Math-7B. In contrast, process-level RL, which provides guidance for each intermediate verification and correction step, may be effective for models with lower initial capabilities, such as Qwen2-7B-Instruct. As shown in Figure \ref{fig:veri_correct_exp}, process-level RL can notably enhance 
the verification and correction abilities of Qwen2-7B-\model-BI.

\begin{table}[!t]
\centering
\scalebox{0.55}{
\begin{tabular}{lcccc}
\toprule[1.5pt]
\textbf{Model} & \textbf{FOLIO} & \textbf{\makecell[c]{CRUX-\\Eval}} &  \textbf{\makecell[c]{Strategy-\\QA}}& \textbf{\makecell[c]{MMLUPro-\\ STEM}} \\ \midrule
Qwen2.5-Math-72B-Instruct &69.5 &68.6 &94.3 &66.0\\
Llama-3.1-70B-Instruct$^*$   & 65.0  & 59.6 & 88.8&61.7  \\
OpenMath2-Llama3.1-70B$^*$   & 68.5  & 35.1 & 95.6 &55.0 \\
QwQ-32B-Preview$^*$          & 84.2  & 65.2 & 88.2 & 71.9 \\
\midrule
Eurus-2-7B-PRIME         &56.7  &\underline{50.0} &79.0  &\bfseries53.7 \\
Qwen2.5-Math-7B-Instruct         & \bfseries61.6 &28.0 & 81.2 &44.7  \\
Qwen2.5-Math-7B           & 37.9&40.8&61.1&46.0\\
 \textbf{Qwen2.5-Math-7B-\model-BI (\textit{ours})}&\underline{58.1} &48.0 &\underline{88.7}&49.8\\
\textbf{Qwen2.5-Math-7B-\model-ORL (\textit{ours})} &\bfseries61.6&\bfseries50.9&\bfseries90.8&\underline{50.0}\\

\bottomrule[1.5pt]
\end{tabular}
}
\caption{Performance of the proposed method and the baseline methods on 4 cross-domain tasks. The results with $^*$ are reported by \citet{shen2025satori}.}
\label{tab:transfer-results}
\vspace{-0.3cm}
\end{table}
\subsection{Generalizing to Cross-domain Tasks}
Despite training on math reasoning tasks, we found that the learned self-verifying and self-correcting capability can also generalize to out-of-distribution general domains. In Table \ref{tab:transfer-results}, we evaluate the SFT model and the outcome-level RL model based on Qwen2.5-Math-7B on four cross-domain tasks: 
FOLIO \cite{han2022folio} on logical reasoning, CRUXEval \cite{gu2024cruxeval} on code reasoning, StrategyQA \cite{geva2021strategyqa} on multi-hop reasoning and MMLUPro-STEM on multi-task complex understanding \cite{wang2024mmlu, shen2025satori}, with details of these datasets provided in Appendix \S\ref{ap:datasets}.
The results show that after learning to self-verify and self-correct, the proposed method effectively boosts the base model's performance across all tasks 
and achieves comparative results to the baseline models. These findings indicate that the learned self-verifying and self-correcting capabilities are general thinking skills, which can also benefit reasoning in general domains. Additionally, we expect that the performance in specific domains can be further improved by applying \modelx training on domain data with minimal reward model requirements (e.g., rule-based or LLM-as-a-judge).
For better illustration, we show cases on how the trained models perform self-verifying and self-correcting on general tasks in Appendix \S\ref{ap:case}.

\subsection{Analyzing Self-verification and Self-correction Abilities}
In this section, we conduct analytical experiments on the models' self-verification and self-correction capabilities from various perspectives.

\subsubsection{Problem-solving v.s. Confirmative Verification}\label{sec:veri_method_exp}
We first compare the Problem-solving and  Confirmative Verification methods described in \S \ref{sec:method_veri}. In Table \ref{tab:veri}, we present the verification results of different methods on the Math500 test set. We report the overall verification accuracy, as well as the initial verification accuracy when the initial answer is correct ($V_{golden}(s_0) = \texttt{correct}$) and incorrect ($V_{golden}(s_0) = \texttt{incorrect}$), respectively.

\begin{table}[h]
\centering
\renewcommand\arraystretch{1.9}
\small
\scalebox{0.64}{
\begin{tabular}{|l|l|c|c|c|}
\hline
\multirow{2}*{\textbf{\makecell[l]{Base Model}}}&\multirow{2}*{\textbf{\makecell[l]{Methods}}} &
\multirow{2}*{\makecell[c]{\bfseries Overall \\ \bfseries Verification\\ \bfseries   Acc.}} & 
\multicolumn{2}{c|}{\makecell[c]{\bfseries Initial Verification Acc.}}\\
\cline{4-5}
&& & \makecell{$V_{golden}(s_0)$\\$ = \texttt{correct}$} & \makecell{$V_{golden}(s_0)$\\$= \texttt{incorrect}$} \\
\hline
\multirow{2}*{{\makecell[l]{ \textit{Llama3.1-8B-Instruct}}}} & Problem-solving & 80.10 & 87.28 & 66.96 \\
\cline{2-5}
& Confirmative & 65.67 & 77.27 & 78.22 \\
\hline
\multirow{2}*{{\makecell[l]{ \textit{Qwen2-7B-Instruct}}}} & Problem-solving & 73.28 & 90.24 & 67.37 \\
\cline{2-5}
& Confirmative & 58.31 & 76.16 & 70.05 \\
\hline
\multirow{2}*{{\makecell[l]{ \textit{Qwen2.5-Math-7B}}}} & Problem-solving & 77.25 & 91.21 & 56.67 \\
\cline{2-5}
& Confirmative & 61.58 & 82.80 & 68.04 \\

\hline

\end{tabular}}
\caption{Comparison of problem-solving and confirmative verification.}
\vspace{-0.2cm}
\label{tab:veri}
\end{table}

We observe from the table that: 
(1) Generally, problem-solving verification achieves superior overall accuracy compared to confirmative verification. This result is intuitive, as existing models are trained for problem-solving, and recent studies have highlighted the difficulty of existing LLMs in performing reverse thinking \cite{berglund2023reversal,chen2024reverse}. During data collection, we also found that existing models tend to verify through problem-solving, even when prompted to verify without re-solving (see Table \ref{tab:veri_follow} in Appendix \S \ref{ap:veri_implement}). 
(2) In practice, accuracy alone does not fully reflect the validity of a method. For example, when answer accuracy is sufficiently high, predicting all answers as correct will naturally lead to high verification accuracy, but this is not a desired behavior. By further examining the initial verification accuracy for both correct and incorrect answers, we found that  problem-solving verification exhibits a notable bias toward predicting answers as correct, while the predictions from confirmative verification are more balanced. We deduce that this bias arises might be because problem-solving verification is more heavily influenced by the preceding solution, aligning with previous studies showing that LLMs struggle to identify their own errors \cite{huang2023large,tyen2023llms}. 
In contrast, confirmative verification performs verification from different perspectives, making it less influenced by the LLMs' preceding solution.

In all experiments, we used confirmative verification for behavior initialization.

\subsubsection{Boosting Self-verifying and Self-correcting with RL}
\label{sec:RL_exp}
In this experiment, we investigate the effect of RL training on the models' self-verifying and self-correcting capabilities.

We assess self-verification using the following metrics:
(1) \textbf{\textit{Verification Accuracy}}: The overall accuracy of verification predictions, as described in \S \ref{sec:veri_method_exp}.
(2) \textbf{\textit{Error Recall}}: The recall of verification when the preceding answers are incorrect.
(3) \textbf{\textit{Correct Precision}}: The precision of verification when it predicts the answers as correct.
Both Error Recall and Correct Precision \ul{directly affect the final answer accuracy}: if verification fails to detect an incorrect answer, or if it incorrectly predicts an answer as correct, the final answer will be wrong.

For self-correction, we use the following metrics: 
(1) \textbf{\textit{Incorrect to Correct Rate}}: the rate at which the model successfully corrects an incorrect initial answer to a correct final answer. 
(2) \textbf{\textit{Correct to Incorrect Rate}}: the rate at which the model incorrectly changes a correct initial answer to an incorrect final answer. 
We provide the formal definitions of the metrics used in Appendix \S\ref{ap:metric}.

\begin{figure}[t]
\centering
\subfloat{\includegraphics[width=1.0\linewidth]{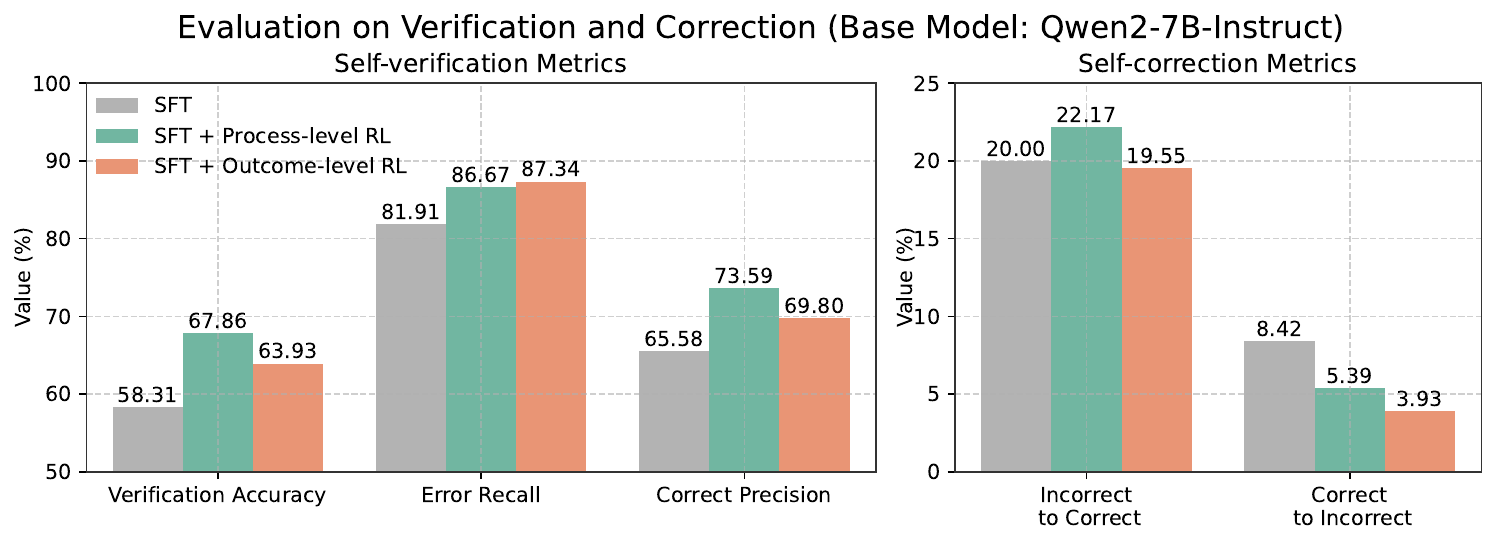}}\\
\subfloat{\includegraphics[width=1.0\linewidth]{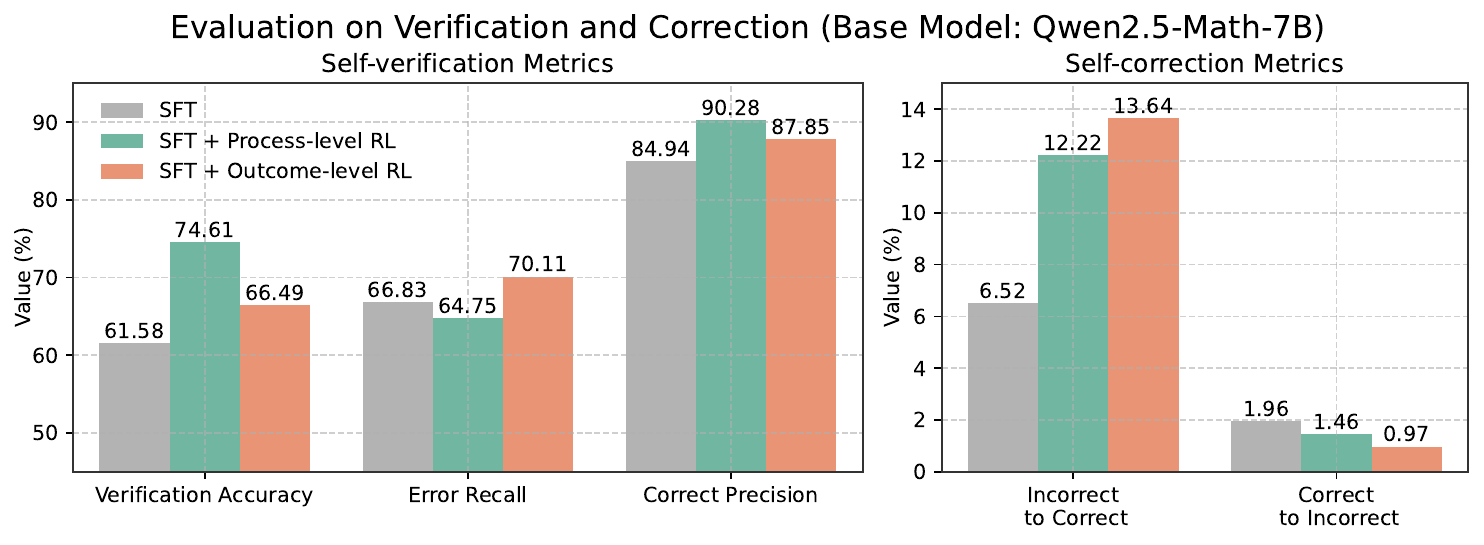}}
\vspace{-0.1cm}
    \caption{Evaluation on verification and correction.}
    \label{fig:veri_correct_exp}
    \vspace{-0.3cm}
\end{figure}

In Figure \ref{fig:veri_correct_exp}, we present the results of the behavior-initialized model (SFT) and different RL models obtained from Qwen2.5-Math-7B. We observe that: (1) Both RL methods effectively enhance self-verification accuracy. The process-level RL shows larger improvement on accuracy, while the outcome-level RL consistently improves Error Recall and Correct Precision. This might be because process-level supervision indiscriminately promotes verification accuracy in intermediate steps, while outcome-level supervision allows the policy model to explore freely in intermediate steps and only boosts the final answer accuracy, thus mainly enhancing Error Recall and Correct Precision (which directly relate to final answer accuracy).
(2) Both RL methods can successfully enhance the models' self-correction capability. Notably, the model's ability to correct incorrect answers is significantly improved after RL training. The rate of model mistakenly altering correct answers is also notably reduced. This comparison demonstrates that \modelx can substantially enhance the validity of models' self-correction ability.

\begin{figure}[h]
\centering
\subfloat{\includegraphics[width=1.0\linewidth]{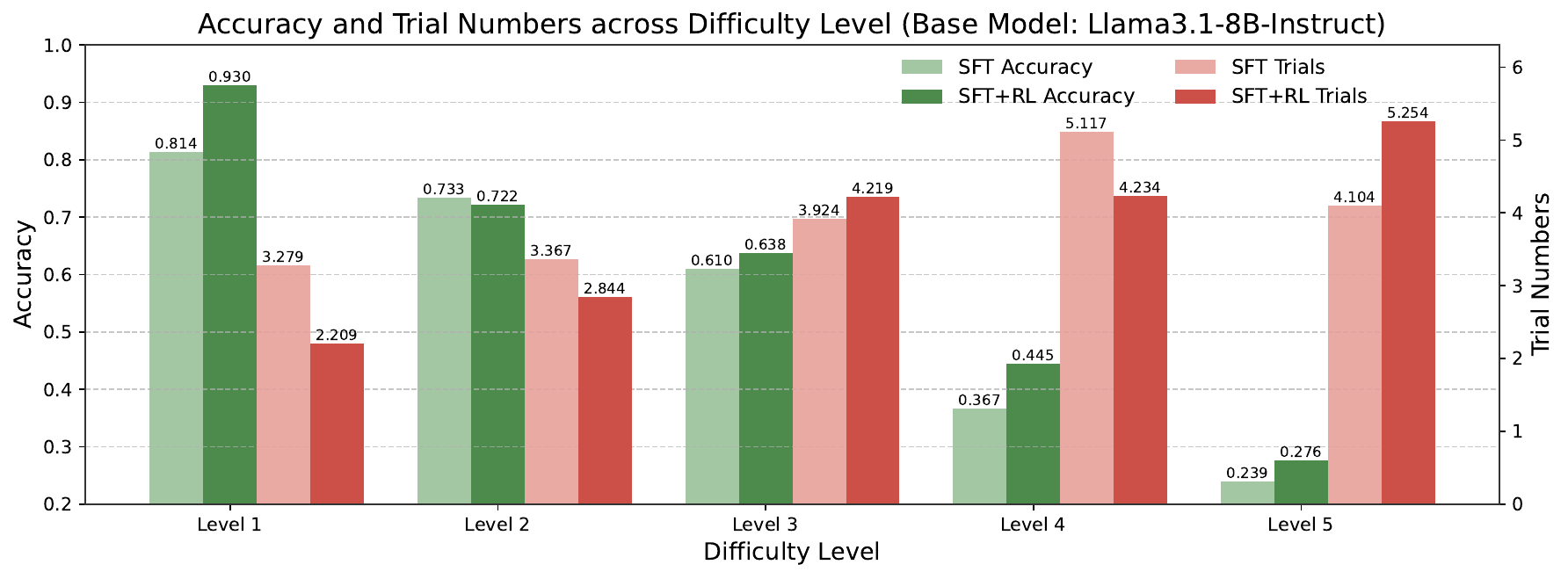}}\\
\subfloat{\includegraphics[width=1.0\linewidth]{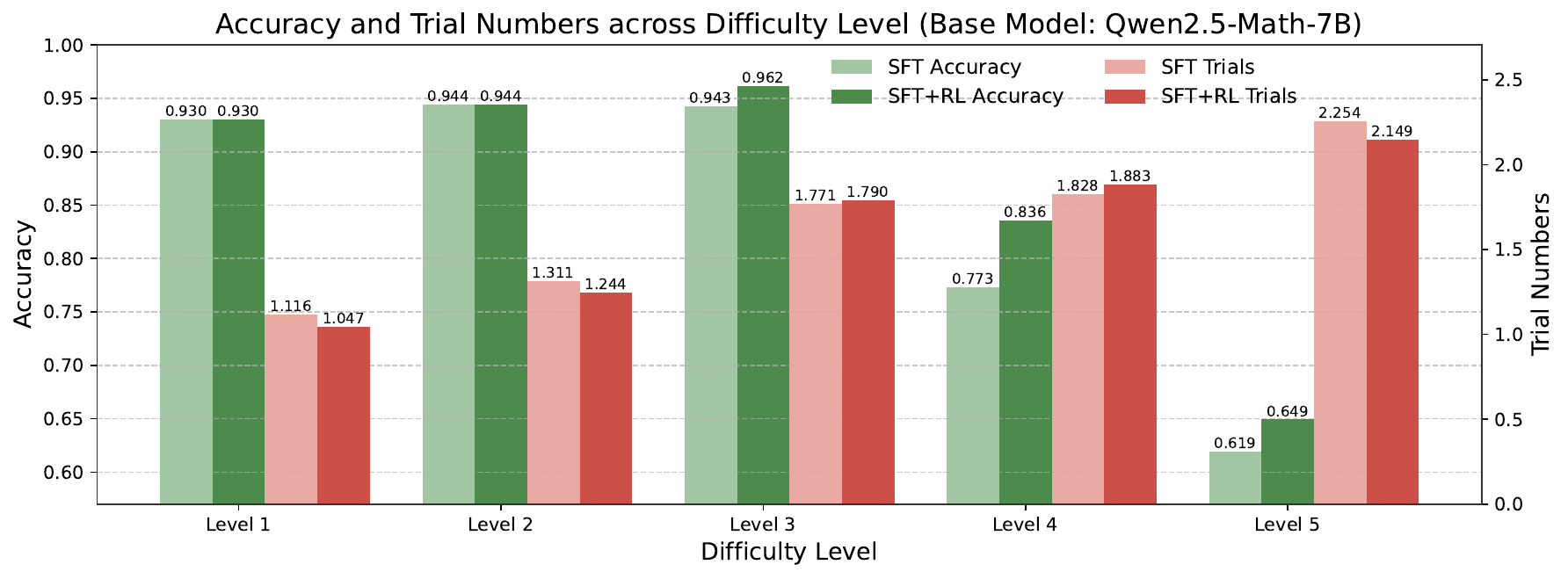}}
\vspace{-0.1cm}
    \caption{The accuracy and average trial number of different models across difficulty levels. Evaluated on MATH500 test set.}
    \label{fig:level_accuracy}
    \vspace{-0.3cm}
\end{figure}

\subsubsection{ Improvement across Difficulty Levels}
To further illustrate the effect of \modelx training, Figure \ref{fig:level_accuracy} shows the answer accuracy and average number of trials (i.e., the average value of "$K$" across all $y=(s_1,v_1,\cdots,s_K,v_K)$ under each difficulty level) for the SFT and SFT+RL models. We observe that:
(1) By learning to self-verify and self-correct during reasoning, the models learn to dynamically allocate test-time effort. For easier problems, the models can reach a confident answer with fewer trials, while for more difficult problems, they require more trials to achieve a confident answer.
(2) RL further improves test-time effort allocation, particularly for less capable model (e.g., Llama3.1-8B-Instruct).
(3) After RL training, the answer accuracy for more difficult problems is notably improved, demonstrating the effectiveness of the self-verifying and self-correcting paradigm in enhancing the models' reasoning abilities.

\begin{table*}[t]
	\small 
	\centering
	\resizebox{0.95\textwidth}{!}{
		\begin{tabular}%
			{@{\hskip0pt}l@{\hskip6pt}c@{\hskip6pt}c@{\hskip6pt}c@{\hskip6pt}c@{\hskip6pt}c@{\hskip6pt}c@{\hskip6pt}c@{\hskip6pt}c@{\hskip6pt}c@{\hskip0pt}}
			\toprule[1.5pt]
			&  \multicolumn{7}{c}{\bf Datasets} & \multirow{3}{*}{\makecell{\textbf{\phantom{xx}Average\phantom{xx}}}} \\
			\cmidrule{2-8} 
			\bfseries Model&\makecell{MATH\\500} &\makecell{AIME\\2024} & \makecell{AMC\\2023}& \makecell{College\\Math}& \makecell{Olympiad \\Bench}&GSM8K&\makecell{GaokaoEn\\2023}\\
		\midrule[1pt]
		\multicolumn{9}{l}{\textit{General Model: Qwen2-7B-Instruct}}\\
		Qwen2-7B-Instruct &51.2 &3.3&30.0 &18.2 &19.1&86.4&39.0&35.3 \\
      	\textbf{Qwen2-7B-\model-BI (\textit{ours})}&61.2 &3.3&27.5&\bfseries 41.1&\bfseries 27.1&87.4&49.1&42.4 \\ \textbf{Qwen2-7B-\model-PRL (\textit{ours})}&\bfseries 65.4&\underline{6.7}&35.0& 36.7&\underline{27.0}&\textbf{89.0}&\underline{49.9}&\underline{44.2}\\ \textbf{Qwen2-7B-\model-ORL (\textit{ours})}&\underline{64.8}&3.3&\bfseries 42.5&34.7 &26.2&86.4&\bfseries 50.9&{44.1}\\
            \textbf{Qwen2-7B--Instruct-\model-PRL-offline (\textit{ours})}&61.6&\bfseries10.0&32.5&40.2&26.5&\underline{87.6}&50.4&44.1 \\
        \textbf{Qwen2-7B-Instruct-\model-ORL-offline (\textit{ours})}&61.0&\underline{6.7}&\underline{37.5}&\underline{40.5}&27.3&87.4&49.6&\bfseries44.3 \\
		\midrule[1pt]
		\multicolumn{9}{l}{\textit{Math-Specialized Model: Qwen2.5-Math-7B}}\\
		Qwen2.5-Math-7B & 51.0 &16.7 &45.0 &21.5  &16.7&58.3&39.7&35.6 \\
		 \textbf{Qwen2.5-Math-7B-\model-BI (\textit{ours})}&81.6 &\underline{23.3} &60.0 & 43.9 &44.4&91.9&70.1&59.3 \\ \textbf{Qwen2.5-Math-7B-\model-PRL (\textit{ours})}&\underline{83.4} &\bfseries{26.7}&\underline{70.0}&43.8&\underline{46.4} &\bfseries{93.2}&\underline{70.4}&\underline{62.0} \\
    \textbf{Qwen2.5-Math-7B-\model-ORL (\textit{ours})}&\bfseries 84.4 &23.3&\bfseries 77.5&43.8&44.9 &\underline{92.9}&{70.1}&\bfseries62.4\\
        \textbf{Qwen2.5-Math-7B-\model-PRL-offline (\textit{ours})}&\underline{83.4} &\underline{23.3}&62.5&\bfseries50.0&\bfseries46.7 &\underline{92.9}&\bfseries 72.2&{61.6} \\
                \textbf{Qwen2.5-Math-7B-\model-ORL-offline (\textit{ours})}&{82.0} &{20.0}&67.5&\underline{49.8}&{45.8} &{92.6}&\underline{70.4}&{61.2} \\
				\midrule[1pt]
\end{tabular}}
\caption{Comparison of \modelx using online and offline RL training.}
\label{tab:offline_result}
\vspace{-0.2cm}
\end{table*}

\subsection{Exploring Offline RL}
As described in \S\ref{sec:offline_intro}, we explore offline RL as a more efficient alternative to online RL training, given the effectiveness of offline RL has been demonstrated in recent studies \cite{baheti2023leftover,cheng2025self,wang2024offline}. 

Table \ref{tab:offline_result} presents the results of offline RL with process-level and outcome-level supervision, compared to online RL. We can observe that: 
(1) Different from online RL, process-level supervision outperforms outcome-level supervision in offline RL training. This interesting phenomenon may be due to: 
a) Outcome-level RL, which excels at allowing models to freely explore dynamic trajectories, is more suitable for on-the-fly sampling during online parameter updating. 
b) In contrast, process-level RL, which requires accurate baseline estimation for intermediate steps, benefits from offline trajectory sampling, which can provide more accurate baseline estimates with larger scale data sampling. 
(2) Offline RL consistently improves performance over the behavior-initialized models across most benchmarks and achieves comparable results to online RL. These results highlight the potential of offline RL as a more efficient alternative for enhancing LLMs' deep reasoning.

\section{Related Work}
\subsection{Scaling Test-time Compute}
Scaling test-time compute recently garners wide attention in LLM reasoning \cite{snell2024scalingllmtesttimecompute,wu2024empirical,brown2024large}. Existing studies have explored various methods for scaling up test-time compute, including: (1) \textbf{\textit{Aggregation-based methods}} that samples multiple responses for each question and obtains the final answer with self-consistency \cite{selfconsistency} or by selecting best-of-N answer using a verifier or reward model \cite{mathshepherd,zhang2024generative,lightman2023lets,havrilla2024glore}; (2) \textit{\textbf{Search-based methods}} that apply search algorithms such as Monte Carlo Tree Search \cite{alphallm,wang2024qimprovingmultistepreasoning,restmcts,qi2024mutual}, beam search \cite{snell2024scalingllmtesttimecompute}, or other effective algorithms \cite{feng2023alphazero,yao2023tree} to search for correct trajectories; (3) \textit{\textbf{Iterative-refine-based  methods}} that iteratively improve test performance through self-refinement \cite{selfrefine,shinn2024reflexion,chen2024magicore,chen2025sets}. Recently, there has been a growing focus on training LLMs to perform test-time search on their own, typically by conducting longer and deeper thinking \cite{o1,guo2025deepseek}. 
These test-time scaling efforts not only directly benefit LLM reasoning, but can also be integrated back into training time, enabling iterative improvement for LLM reasoning \cite{qin2024o1,feng2023alphazero,snell2024scalingllmtesttimecompute,luong2024reft}. 
In this work, we also present an efficient framework for training LLMs to perform effective test-time scaling through self-verification and self-correction iterations. This approach is achieved without extensive efforts, and the performance of \modelx can also be consistently promoted via iterative training.

\subsection{Self-verification and Self-correction}
Enabling LLMs to perform effective self-verification and self-correction is a promising solution for achieving robust reasoning for LLMs \cite{madaan2024self,shinn2023reflexion,paul2023refiner,lightman2023let}, and these abilities are also critical for performing deep reasoning.
Previous studies have shown that direct prompting of LLMs for self-verification or self-correction is suboptimal in most scenarios \cite{huang2023large,tyen2023llms,ma2024large,zhang2024understanding}. As a result, recent studies have explored various approaches to enhance these capabilities during post-training \cite{saunders2022self,rosset2024direct,kumar2024training}. These methods highlight the potential of using human-annotated or LLM-generated data to equip LLMs with self-verification or self-correction capabilities \cite{zhang2024small,jiang2024towards}, while also indicating that behavior imitation via supervised fine-tuning alone is insufficient for achieving valid self-verification or self-correction  \cite{kumar2024training,qu2025recursive,kamoi2024can}. In this work, we propose effective methods to enhance LLMs' self-verification and self-correction abilities through principled imitation data construction and RL training, and demonstrate the effectiveness of our approach with in-depth analysis.

\subsection{RL for LLM Reasoning}
Reinforcement learning has proven effective in enhancing LLM performance across various tasks \cite{ziegler2019fine,stiennon2020learning,bai2022training,ouyang2022training,setlur2025rl}. In LLM reasoning, previous studies typically employ RL in an actor-critic framework \cite{lightman2024lets,tajwar2024preference,havrilla2024teaching}, and research on developing accurate reward models for RL training has been a long-standing focus, particularly in reward modeling for Process-level RL \cite{lightman2024lets,setlur2024rewarding,setlur2025rl,luo2024improve}. Recently, several studies have demonstrate that simplified reward modeling and advantage estimation \cite{ahmadian2024back,deepseekmath,team2025kimi,guo2025deepseek} in RL training can also effectively enhance LLM reasoning. Recent advances in improving LLMs' deep thinking \cite{guo2025deepseek,team2025kimi} further highlight the effectiveness of utilizing unhackable rewards \cite{gao2023scaling,everitt2021reward} to consistently enhance LLM reasoning.
In this work, we also show that simplified advantage estimation and RL framework enable effective improvements on LLM reasoning. Additionally, we conducted an analysis on process-level RL, outcome-level RL and offline RL, providing insights for future work in RL for LLM reasoning.

\section{Conclusion}
In this work, we propose \model, an efficient framework for enhancing LLM reasoning by teaching LLMs to iteratively self-verify and self-correct during reasoning. 
We introduce a principled approach for behavior initialization, and explore both outcome-level and process-level RL to further strengthen the models' thinking abilities.
Experimental results across three different base models on seven math reasoning benchmarks demonstrate that \modelx significantly enhances LLM reasoning with minimal resource requirements. 
Since self-verification and self-correction are two crucial abilities for LLMs' deep reasoning, \modelx offers an interpretable framework for understanding how SFT and RL enhance LLMs' deep reasoning. It also offers insights into the selection of RL strategies for enhancing LLMs' long-CoT reasoning.

\bibliography{custom}

\appendix
\label{sec:appendix}

\clearpage
\section{Implementation Details}
\subsection{Verification Processing and SFT Data Construction}
\label{ap:veri_implement}
Given the responses sampled from the original LLM policy, we prompt frontier LLMs for initial verifications. In order to construct more valid verification, we force the LLMs to ``verify without re-solving the problem'' and filter out invalid verifications during data processing. We found that despite being instructed to "verify without re-solving the problem", most existing LLMs still biased to solve the problem again, as shown in Table \ref{tab:veri_follow}. Finally, we collected the verification data by querying gpt-4-preview-1106\footnote{\url{https://openai.com/api/}}
, which shows strong instruction-following ability to "verify without re-solving the problem" and can perform plausible verification such as adopting reverse thinking, inductive reasoning and other methods.

For these collected prompts, we refine the remaining verifications using gpt-4o to improve fluency and clarity. During this refinement, we instruct gpt-4o to append a conclusion at the end of each verification based on its stance—for example: ``Therefore, the answer is correct/incorrect/cannot verify.'' Finally, we discard any verifications where the judgment does not align with the actual correctness of the answer. The prompts we used during the whole process are provided in Appendix \S\ref{ap:prompts}.

\begin{figure*}[t]
	\centering
	\includegraphics[width=1.0\textwidth]{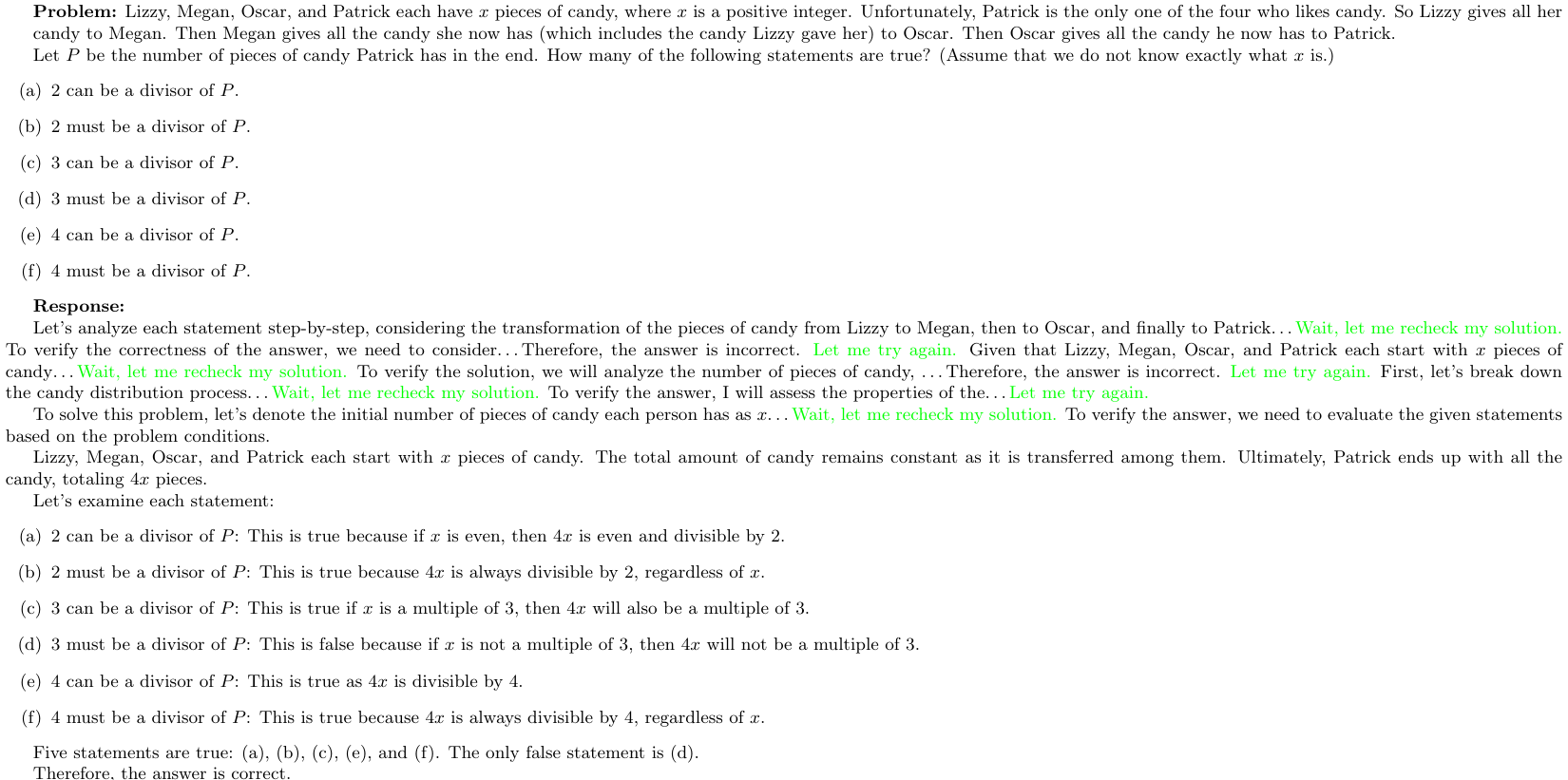}
	\caption{SFT data example.}
	\label{fig:sft_case}
\end{figure*}

With the refined and filtered verifications, we construct the SFT data as follows. For each problem, we determine the number of answer attempts required to eventually obtain a correct answer based on the accuracy from the initial sampling. The lower the accuracy, the more rounds of responses are generated. In our implementation, we categorize all problems into four difficulty levels and construct answer sequences with 1, 2, 3, or 4 rounds, according to descending accuracy. Then, after an incorrect answer, we append ``Wait, let me recheck my solution'' along with the corresponding verification. If that answer is not the final attempt, we further append ``Let me try again.'' We ensure that the last answer in the sequence is correct. Additionally, we ensure that the answers in each round for a given problem are distinct. Figure \ref{fig:sft_case} is an example of SFT data constructed with 4 rounds of responses.

\subsection{Baseline Details}\label{ap:baseline}
\subsubsection{Baseline Implementations}

In Table \ref{tab:mainresults}, the reported results for Frontier LLMs and Top-tier Open-source Reasoning LLMs are sourced from the original reports and \citet{guan2025rstar}. We evaluate Llama-3.1-8B-Instruct \cite{llama3.1}, Qwen2-7B-Instruct \cite{qwen2}, Qwen2.5-Math-7B, Qwen2.5-Math-7B-Instruct and Qwen2.5-Math-72B-Instruct\cite{qwen2.5} using the same process described in Section \S \ref{sec:e_metric}. For Eurus-7B-PRIME \cite{cui2025process}, rStar-Math-7B \cite{guan2025rstar}, and Qwen2.5-7B-SimpleRL \cite{zeng2025simplerl}, we report results directly from the original papers.

In Table \ref{tab:transfer-results}, the results for Llama-3.1-70B-Instruct and QwQ-32B-Preview are taken from \citet{shen2025satori}. For the remaining baselines, we follow the official evaluation protocol of the dataset project\footnote{%
  \begin{tabular}[t]{@{}l@{}}
    \href{https://github.com/Yale-LILY/FOLIO}{https://github.com/Yale-LILY/FOLIO}\\[0.5ex]
    \href{https://github.com/facebookresearch/cruxeval}{https://github.com/facebookresearch/cruxeval}\\[0.5ex]
    \href{https://github.com/eladsegal/strategyqa}{https://github.com/eladsegal/strategyqa}\\[0.5ex]
    \href{https://github.com/TIGER-AI-Lab/MMLU-Pro}{https://github.com/TIGER-AI-Lab/MMLU-Pro}
  \end{tabular}
}.
\subsubsection{Baseline License}

In this work, we utilize the Llama-3.1-8B-Instruct model, whose license can be reviewed at \url{https://huggingface.co/meta-llama/Llama-3.1-8B-Instruct/blob/main/LICENSE}. In addition, the models Qwen2-7B-Instruct, Qwen2.5-Math-7B, Eurus-2-7B-PRIME, and project vLLM are distributed under the Apache License 2.0. We gratefully acknowledge the contributions of the open-source community and strictly adhere to the terms of the respective licenses.

\subsubsection{Baseline SFT Data Construction}
\label{ap:sft_base}
\paragraph{Original Solution SFT Data}
In this setting, we use the solution from the original dataset as sft data. To ensure a fair comparison, we maintain the same training data volume as our behavior initialization approaches.
\paragraph{Long CoT SFT Data}

We also introduce a baseline by fine-tuning on Long CoT responses generated by QwQ-32B-Preview \cite{qwq-32b-preview}. Specifically, we instruct QwQ to generate responses to given problems and filter out those with incorrect answers. The remaining high-quality responses are then used for supervised fine-tuning. Importantly, we ensure that the total training data volume remains consistent with that used in our behavior initialization approach. The prompt we use for QwQ is provided in Appendix \S\ref{ap:prompts}.

\subsection{Prompts}
\label{ap:prompts}
The prompts we use in all experiments are as follows:
\begin{lstlisting}[title={Sampling Responses During Training/Inference}]
Please reason step by step, and put your final answer within 
\boxed{}. 
Problem: {problem} 
\end{lstlisting}

\begin{lstlisting}[title={Verification Refinement}]
You are a math teacher. I will give you a math problem and an answer. 
Verify the answer's correctness without step-by-step solving. Use alternative verification methods. 
Question: {problem}
Answer: {answer}
Verification:
\end{lstlisting}

\begin{lstlisting}[title={Verification Collection}]
Refine this verification text to read as a natural self-check within a solution. Maintain logical flow and professionalism.
Key Requirements:
1. Avoid phrases like "without solving step-by-step" or "as a math teacher".
2. Treat the answer as your own prior solution.
3. Conclude with EXACTLY one of:
Therefore, the answer is correct.
Therefore, the answer is incorrect.
Therefore, the answer cannot be verified.
Original text: {verification}
\end{lstlisting}

\section{Detailed Experiment Settings}
\begin{table}[]
\resizebox{\columnwidth}{!}{%
\begin{tabular}{@{}lc@{}}
\toprule[1.5pt]
 \multicolumn{2}{l}{\textit{Without Asking for Confirmative Verification}}\\
Model & Confirmative out of 100\\ \midrule[1pt]
GPT-4o     &     26                           \\
GPT-4-Preview-1106  &    32                             \\
QwQ-32B-preview      &               37               \\
Llama-3.1-70B-Instruct      &       28        \\

      \bottomrule
      \toprule[1.5pt]
   \multicolumn{2}{l}{\textit{Asking for Confirmative Verification}}\\
Model & Confirmative out of 100 \\ \midrule[1pt]
GPT-4o     &     44                          \\
GPT-4-Preview-1106  &           \textbf{61}                                        \\
QwQ-32B-preview      &                       58                                      \\
Llama-3.1-70B-Instruct      &          50                  \\
      \bottomrule
\end{tabular}%
}
\caption{}
\label{tab:veri_follow}
\end{table}
\subsection{Datasets}
\label{ap:datasets}
Details of each test dataset we used as benchmark are as follows:

\subsubsection{In-domain Datasets}
\textbf{MATH500} \cite{lightman2023lets} offers a streamlined slice of the broader MATH \cite{MATH} dataset, comprising 500 test problems selected through uniform sampling. Despite its smaller scope, it maintains a distribution of topics and difficulty levels that mirrors the larger MATH corpus.

\textbf{GSM8K} \cite{cobbe2021gsm8k} features around 8,500 grade-school math word problems. The dataset focuses on simple arithmetic through early algebra and includes 1,319 distinct tasks in its test set.

\textbf{OlympiadBench} \cite{he2024olympiadbench} collects 8,476 advanced math and physics questions drawn from Olympiad contexts, with some originating from the Chinese college entrance exam. We use the subset of 674 text-only competition questions, providing open-ended math challenges.

\textbf{AMC2023} \cite{amc} and \textbf{AIME} \cite{aime} each supply a set of challenging exam-style problems: 40 questions from AMC 2023 and 30 from AIME 2024, all in text-only format.

\textbf{CollegeMath} \cite{tang2024mathscale} is a dataset targeting advanced college-level mathematics, drawn from nine textbooks spanning seven major fields—algebra, pre-calculus, calculus, vector calculus, probability, linear algebra, and differential equations. The final collection comprises 1,281 training examples and 2,818 test examples.

\textbf{Gaokao2023en} \cite{liao2024mario} is a dataset consisting of 385 mathematics problems sourced from the 2023 Chinese higher education entrance examination, which have been professionally translated into English. 

\subsubsection{Cross-domain Datasets}
\textbf{FOLIO} \cite{han2022folio} is meticulously annotated to assess intricate logical reasoning in natural language. It pairs 1,430 conclusions with 487 sets of premises—each verified using first-order logic (FOL)—and contains 203 unique problems in its test portion.

\textbf{CRUXEval} \cite{gu2024cruxeval} tests code comprehension and reasoning through 800 concise Python functions (spanning 3–13 lines). Each function is accompanied by one or more input-output examples. The goal is to predict the correct outputs given the function body and a specific input. The test partition encompasses all 800 problems.

\textbf{StrategyQA} \cite{geva2021strategyqa} targets multi-hop reasoning questions where the necessary intermediate steps are not explicit. Each of its 2,780 items includes a strategic query, a breakdown of the reasoning steps, and supporting evidence drawn from Wikipedia.

\textbf{MMLUProSTEM} is extracted from \textbf{MMLU-Pro} \cite{wang2024mmlu}. Following Satori \cite{shen2025satori}, we conduct evaluations on six STEM subsets—physics, chemistry, computer science, engineering, biology, and economics.

\subsection{Hyperparameters Setting}
\label{ap:hyper}
\begin{table*}[]
\centering
\resizebox{\textwidth}{!}{%
\begin{tabular}{@{}lcccccc@{}}
\toprule[1.5pt]
 & Model & Learning Rate  & Batch Size & KL Coefficient&Max Length & Training Epochs \\ 
\midrule[1pt]
& Llama-3.1-8B-Instruct & 5e-6  & 32 & 0.1&8000& 3\\
& Qwen2-7B-Instruct & 5e-6 & 32 & 0.1 &6000& 3 \\
& Qwen2.5-Math-7B & 5e-6  & 32 & 0.01&8000& 3 \\ 
\bottomrule[1.5pt]
\end{tabular}%
}
\caption{Model Training Hyperparameter Settings (SFT)}
\label{tab:hyper_sft}
\end{table*}

\begin{table*}[]
\centering
\resizebox{\textwidth}{!}{%
\begin{tabular}{@{}lccccccccc@{}}
\toprule[1.5pt]
 & Model & Learning Rate  & \makecell[c]{Training\\Batch Size} & \makecell[c]{Forward\\Batch Size} & KL Coefficient&Max Length & \makecell[c]{Sampling\\Temperature} &Clip Range &Training Steps \\ 
\midrule[1pt]
& Llama-3.1 &5e-7  & 64& 256 & 0.05&8000& 0.7&0.2&500\\
& Qwen2-7B-Instruct & 5e-7&  64& 256 & 0.05 &6000&0.7 &0.2&500\\\
& Qwen2.5-Math-7B & 5e-7 & 64& 256 & 0.01&8000&0.7 &0.2&500 \\ 
\bottomrule[1.5pt]
\end{tabular}%
}
\caption{Model Training Hyperparameter Settings (RL)}
\label{tab:hyper_rl}
\end{table*}
During behavior initialization with SFT, we use a batch size of \textit{32} and adopt a learning rate of \textit{5e-6}. We set the maximum sequence length \textit{8000} to accommodate long responses and verifications. To balance stability and convergence during training, we add a KL punishment to the training loss, and the KL coefficient is set to \textit{0.1}. 

During reinforcement learning, 
for each training batch, we use a training batch size of 64, and sample $n$ responses for each question in a batch, resulting a forward batch size of $64n$. For each forward batch, we update the model for $n$ step with the training batch size 64. Specifically, for both process-level and outcome-level RL, we adopt $n=4$ (i.e., for RLOO, the sample number is also $4$). More hyperparameters of the RL training are presented in Table \ref{tab:hyper_rl}. We use the BF16 model precision in all experiments.

Main hyperparameters used in the experiments are illustrated in Table \ref{tab:hyper_sft} and \ref{tab:hyper_rl}.

\subsection{Experiment Environment}

All experiments are implemented using the PyTorch framework on 32 NVIDIA H20 (96GB) GPUs or 32 NVIDIA A100Pro (40GB) GPUs. Our training code is built upon Hugging Face TRL\footnote{\url{https://github.com/huggingface/trl}}. For inference, we use a single NVIDIA A100 (40GB) GPU with vLLM-0.5.4\footnote{\url{https://github.com/vllm-project/vllm}}. 
We utilize transformers version 4.39.3 for fine-tuning Qwen2-7B-Instruct and Qwen2.5-Math-7B, version 4.44.0 for fine-tuning Llama-3.1-8B, and version 4.46.3 for reinforcement learning. We use PyTorch 2.1.1 across our training pipeline.
Our evaluation code is built upon Qwen Math's evaluation codebase\footnote{\url{https://github.com/QwenLM/Qwen2.5-Math}}.

\section{Metrics Definition}
\label{ap:metric}
We include the formal definition of metrics we use for analyzing self-verification and self-correction behaviors of the post-trained models as follows.

\subsection{Notations}
We first present the main notations used in our formulation in Table \ref{tab:variable_lookup}.

\begin{table}
    \centering
    \scalebox{0.8}{
    \begin{tabular}{c|l}
    \hline
    \textbf{Variable} & \textbf{Description} \\ \hline
    $\pi$ & The policy \\ \hline
    $x$ & Problem instance \\ \hline
    $y$ & \makecell[c]{Series of predefined actions: \\$y = \{a_1, a_2, \ldots, a_n\}$} \\ \hline
    $a_i$ & The $i$-th action in the response $y$, and let  \\ 
    & $Type(a_i) \in \{\texttt{verify}, \texttt{solve}, \texttt{<end>}\}$ \\ \hline

    $s_j$ & $j^{th}$ attempt to solve the problem \\ \hline
    $v_j$ & $j^{th}$ self-verification for the $j^{th}$ attempt \\ \hline
    $Parser(\cdot)$ & $Parser(v_j) \in \{\texttt{correct}, \texttt{incorrect}\}$ \\
     & The text parser to get the self-verification result \\ 
     & indicating the correctness of action $s_j$  \\ \hline
    $V_{golden}(\cdot)$ & $V_{golden}(a_i) \in \{\texttt{correct}, \texttt{incorrect}\}$ \\ \hline
    $R(\cdot)$ & The rule based reward function   \\ 
    & $R(\cdot)\in \{-1, 1\}$   \\

    &
    $
    \small
    R(s_j) = 
    \begin{cases}
    1,& V_{golden}(s_j) = \texttt{correct} \\
    -1, & otherwise \\
    \end{cases}
    $ \\

    &
    $
    \small
    R(v_j) = 
    \begin{cases}
    1,& Parser(v_j) = V_{golden}(s_j) \\
    -1, & otherwise \\
    \end{cases}
    $
    \\ \hline

    \texttt{<end>} & End of action series \\ \hline
    {$\mathbb{I}(\cdot)$} & \makecell[l]{The indicator function, $\mathbb{I}(\cdot)\in\{0,1\}$.\\
    $\mathbb{I}(\cdot)=1$ if the condition inside holds true,\\ and $\mathbb{I}(\cdot)=0$ otherwise.}\\
    \hline
    \end{tabular}}
    \caption{Variable Lookup Table}
    \label{tab:variable_lookup}
\end{table}

\subsection{Self-Verification Metrics}

\subsubsection{Verification Accuracy (VA)}
Verification Accuracy measures how often the verification prediction matches the ground-truth correctness ($N$ is the total number of verifications in the responses to the test set):
\begin{equation}
\resizebox{\linewidth}{!}{
$\text{VA} = \frac{1}{N}\sum_{t=1}^{N} \mathbb{I}\Bigl(\text{Parser}(v_t) = V_{golden}(s_t)\Bigr).$
}
\end{equation}

\subsubsection{Error Recall (ER)}
Error Recall measures the recall of detecting incorrect answers (i.e., the fraction of actually incorrect answers that are successfully identified as incorrect):
\begin{equation}
\label{eq:error_recall}
\resizebox{\linewidth}{!}{$
\text{ER} = \frac{\sum_y\sum_{t=1}^\frac{|y|_a}{2} 
    \mathbb{I}\Bigl(R(s_{t})=-1\Bigr)\,
    \mathbb{I}\Bigl(\text{Parser}(v_t)=\texttt{incorrect}\Bigr)}
  {\sum_y\sum_{t=1}^{\frac{|y|_a}{2}}
    \mathbb{I}\Bigl(R(s_{t})=-1\Bigr)}.
$}
\end{equation}
where $|y|_a$ is the total number of actions in $y$ and $\frac{|y|_a}{2}$ is the total number of attempts to solve the problem ($y=\{a_1, a_2,\cdots,a_{|y|_a}\}=\{s_1, v_1,\cdots,s_\frac{|y|_a}{2},v_\frac{|y|_a}{2} \}$). 

\subsubsection{Correct Precision (CP)}
Correct Precision measures the precision when the verification model predicts an answer to be correct (i.e., among all ``correct'' predictions, how many are truly correct):
\begin{equation}
\resizebox{\linewidth}{!}{$
\text{CP} = \frac{\sum_y\sum_{t=1}^\frac{|y|_a}{2}
    \mathbb{I}\Bigl(\text{Parser}(v_t)=\texttt{correct}\Bigr)\,
    \mathbb{I}\Bigl(R(s_{t})=1\Bigr)}
  {\sum_y\sum_{t=1}^\frac{|y|_a}{2}
    \mathbb{I}\Bigl(\text{Parser}(v_t)=\texttt{correct}\Bigr)}.
$}
\end{equation}

\subsection{Self-Correction Metrics}

\subsubsection{Incorrect to Correct Rate (ICR)}
The rate at which the model successfully corrects an initially incorrect answer ($R(s_1)=-1$) into a correct final answer ($R(s_{T_y})=1$), where $T_y=|y|_a/2$ is the total number of attempts to
solve the problem in each $y$. Formally:
\begin{equation}
\text{ICR} 
= \frac{\sum_y \mathbb{I}\bigl(R(s_1) = -1\bigr)\,\mathbb{I}\bigl(R(s_{T_y}) = 1\bigr)}
       {\sum_y \mathbb{I}\bigl(R(s_1) = -1\bigr)}.
\end{equation}

\subsubsection{Correct to Incorrect Rate (CIR)}
The rate at which the model incorrectly alters an initially correct answer ($R(s_1)=1$) into an incorrect final answer ($R(s_{T_y})=-1$), where $T_y=|y|_a/2$ is the total number of attempts to
solve the problem in each $y$. Formally:
\begin{equation}
\label{eq:correct_to_incorrect}
\text{CIR} 
= \frac{\sum_y \mathbb{I}\bigl(R(s_1) = 1\bigr)\,\mathbb{I}\bigl(R(s_{T_y}) = -1\bigr)}
       {\sum_y \mathbb{I}\bigl(R(s_1) = 1\bigr)}.
\end{equation}

\section{Offline RL Training Details}

In this section, we provide additional details on the offline reinforcement learning training process, including formal definition, ablation studies, and implementation details.

\subsection{Accuracy-Grouped Baseline Definition}
\label{ap:acc_baseline}
To fully leverage the advantages of offline RL, which does not require real-time sampling, we explore more appropriate baseline selection by further grouping trajectories based on problem difficulty. Intuitively, for two trajectories $y^{(1)}$ and $y^{(2)}$ sampled under questions of different difficulty levels, and their corresponding actions $a^{(1)}_t$ and $a^{(2)}_t$ at the same position, even if they share identical reward contexts, their expected returns (baselines) should differ, i.e., the expected return is typically lower for more challenging problems. 

We measure a problem’s difficulty by estimating how often it is solved correctly under the current sampling policy. Concretely, we sample multiple trajectories in parallel for each problem. The fraction of these trajectories that yield a correct final answer serves as the problem’s accuracy. We then discretize this accuracy into separate bins, effectively grouping the problems according to their estimated difficulty. All trajectories belonging to problems within the same accuracy bin form a common subset.

Compared to using direct reward contexts alone, this accuracy-based grouping offers a more robust estimate of expected returns, problems in the same bin share similar success rates. Moreover, unlike a pre-defined difficulty grouping, these bins adjust dynamically as the model’s capabilities evolve. Building on this approach, we propose two accuracy-based baseline estimation methods for offline RL as follows.

\subsubsection{Accuracy-Grouped Baseline With Position Group}
\label{ap:acc_baseline_position}
Within each accuracy bin, we further split actions based on their position in the trajectory. Concretely, we consider all actions occurring at the same step index across trajectories in the same bin to be comparable, and we compute their average return to serve as the baseline. Thus, when we look up the baseline for a particular action at a given step in a trajectory, we use the average return of all actions taken at that same step index in all trajectories belonging to the same accuracy bin.

\subsubsection{Accuracy-Grouped Baseline With Reward Context}
We also propose combining accuracy-based grouping with reward-context grouping. The underlying assumption is that even if two actions share the same immediate reward context, their expected returns can differ if they originate from different difficulty bins. Generally, problems that are harder to solve exhibit lower expected returns. Consequently, we first bin the trajectories by accuracy, then further group them by common reward context. Within each sub-group, we average the returns of all relevant actions to obtain the baseline.

\subsection{Offline RL Implementation Details}
\label{ap:offline_rl_details}
In each iteration of offline RL training, we generate multiple trajectories (e.g., eight) per prompt in parallel. We then apply prompt filtering, rejection sampling, accuracy-based baseline estimation, advantage computation, and policy updates. Implementation details follow.

\subsubsection{Prompt Filtering}
\begin{table*}[t]
\small 
\centering
\resizebox{\textwidth}{!}{
\begin{tabular}%
{@{\hskip0pt}l@{\hskip4pt}c@{\hskip6pt}c@{\hskip6pt}c@{\hskip6pt}c@{\hskip6pt}c@{\hskip6pt}c@{\hskip6pt}c@{\hskip6pt}c@{\hskip6pt}c@{\hskip0pt}}
\toprule[1.5pt]
\textbf{Accuracy Range} & \textbf{Retained Questions} & \textbf{MATH500} & \textbf{AIME2024} & \textbf{AMC2023} & \textbf{College Math} & \textbf{Olympiad Bench} & \textbf{GSM8K} & \textbf{GaokaoEn2023} & \textbf{Average} \\
\midrule
$[0.1-0.7]$ & 1805 & 83.4 & 23.3 & 62.5 & 50.0 & 46.7 & 92.9 & 72.2 & 61.6 \\
$[0.2-0.8]$ & 2516 & 82.6 & 23.3 & 70.0 & 49.8 & 45.3 & 92.4 & 70.1 & 61.9 \\
$[0.3-0.9]$ & 4448 & 81.6 & 23.3 & 70.0 & 49.4 & 44.7 & 92.0 & 68.1 & 61.3 \\
$[0-1]$ & Full & 80.6 & 26.7 & 67.5 & 50.0 & 43.0 & 91.4 & 67.0 & 60.9 \\
\midrule[1pt]
\end{tabular}}
\caption{Comparison of question filtering accuracy selection.}
\label{tab:acc_ablation}
\end{table*}
As we sample multiple trajectories for each prompt, we compute the accuracy of each prompt. We retain prompts whose accuracy falls within a predefined range.

Our ablation study on Qwen2.5-Math-7B shown in Table \ref{tab:acc_ablation} confirms that filtering improves performance. The most stable results are obtained with an accuracy range of \([0.1,0.7]\), suggesting that including moderately difficult samples enhances the model's reasoning capabilities.

\subsubsection{Rejection Sampling}
We discard any trajectory that does not follow the alternation pattern of solution and verification:
$y = (s_1, v_1, \dots, s_k, v_k)$.
Additionally, we remove malformed trajectories such as $y = (s_1, s_2, v_1)$. To mitigate reward hacking due to excessively long outputs, we eliminate trajectories where $R(s_t) = 1$ and $R(v_t) = 1$ at timestep $t$, but further actions are taken at $t+1$. Moreover, we discard trajectories containing more than 20 actions, as excessive action sequences can introduce instability and deviate from expected solution structures.

\subsubsection{Loss Function}
\begin{table*}[t]
\small 
\centering
\resizebox{\textwidth}{!}{
\begin{tabular}%
{@{\hskip0pt}l@{\hskip4pt}c@{\hskip6pt}c@{\hskip6pt}c@{\hskip6pt}c@{\hskip6pt}c@{\hskip6pt}c@{\hskip6pt}c@{\hskip6pt}c@{\hskip6pt}c@{\hskip0pt}}
\toprule[1.5pt]
&  \multicolumn{7}{c}{\bf Datasets} & \multirow{3}{*}{\makecell{\textbf{\phantom{xx}Average\phantom{xx}}}} \\
\cmidrule{2-8} 
\bfseries Baseline Method & \textbf{MATH500} & \textbf{AIME2024} & \textbf{AMC2023} & \textbf{College Math} & \textbf{Olympiad Bench} & \textbf{GSM8K} & \textbf{GaokaoEn2023} &  \\
\midrule[1pt]
\textbf{Based on reward context} & 82.4 & 26.7 & 65.0 & 50.1 & 46.1 & 92.9 & 71.2 & 62.1\\
\textbf{Based on accuracy group with position} & 83.4 & 23.3 & 62.5 & 50.0 & 46.7 & 92.9 & 72.2 & 61.6\\
\textbf{Based on accuracy group with reward context} & 82.4 & 23.3 & 67.5 & 49.3 & 45.8 & 93.3 & 71.2 & 61.8\\
\midrule[1pt]
\end{tabular}}
\caption{The performance of different baselines}
\label{tab:baseline_ablation}
\end{table*}
To determine the best offline baseline method, we conducted ablation studies on Qwen2.5-Math-7B shown in Table \ref{tab:baseline_ablation}. We found that using the accuracy-grouped baseline with an additional division by position provides the most stable results. When computing advantages, we subtract both the baseline and a scaled relative policy term like Equation 5. Notably, we fix $\pi_{\text{ref}}$ as the reference policy instead of being updated at each iteration.

\subsubsection{Training Hyperparameter Settings}
We use a batch size of 64, a maximum learning rate of $5\times10^{-7}$, and a KL penalty coefficient of 0.1. The maximum training sequence length is set to 8192. We apply a warm-up phase of 5 steps and a clipping range parameter of 0.2. We use BF16 model precision in all experiments.

\section{Demo Cases}
\label{ap:case}

\begin{figure*}[t]
	\centering
	\includegraphics[width=1.0\textwidth]{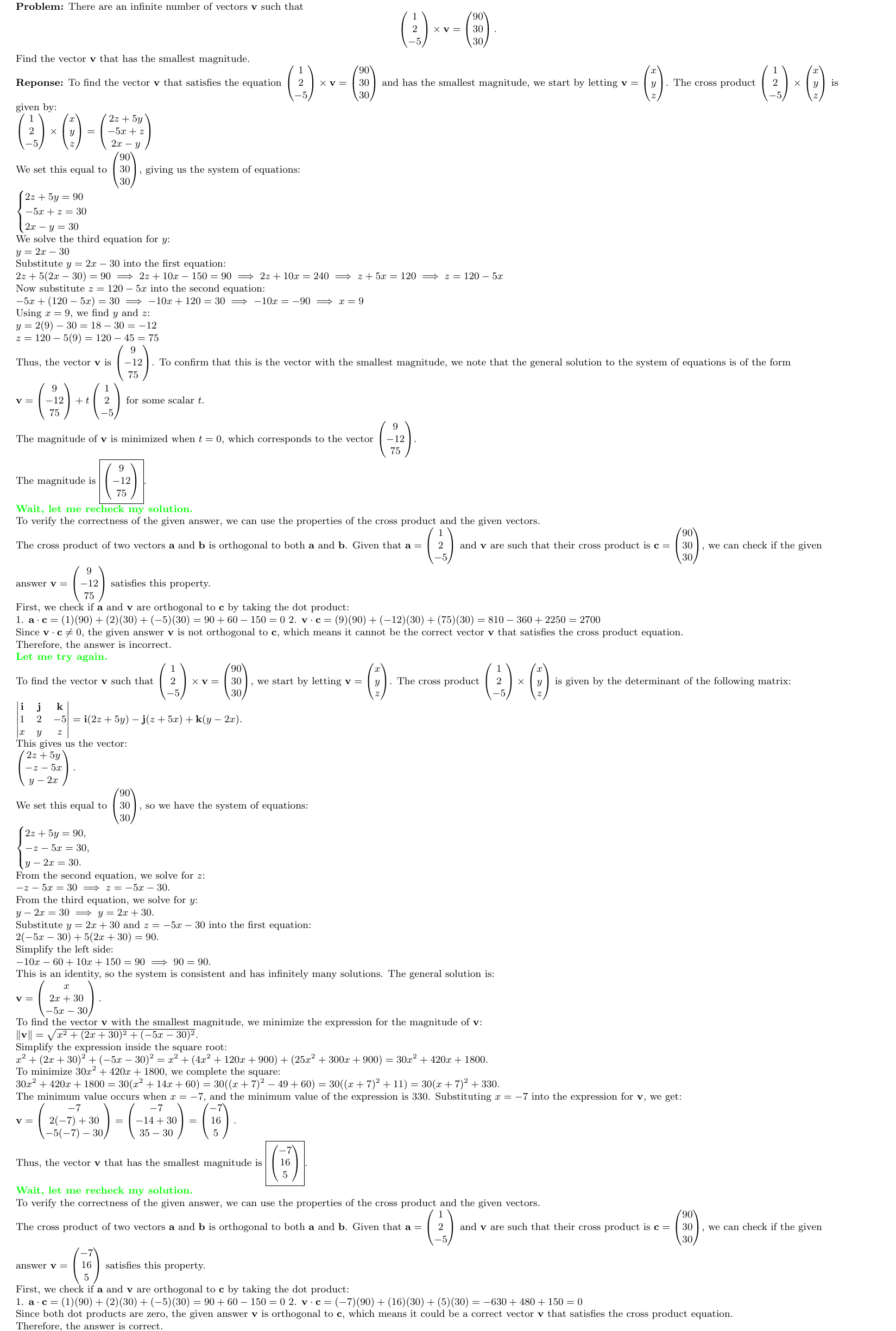}	
	\caption{MATH500 Case.}
	\label{fig:math_case}
\end{figure*}

\begin{figure*}[t]
	\centering
	\includegraphics[width=1.0\textwidth]{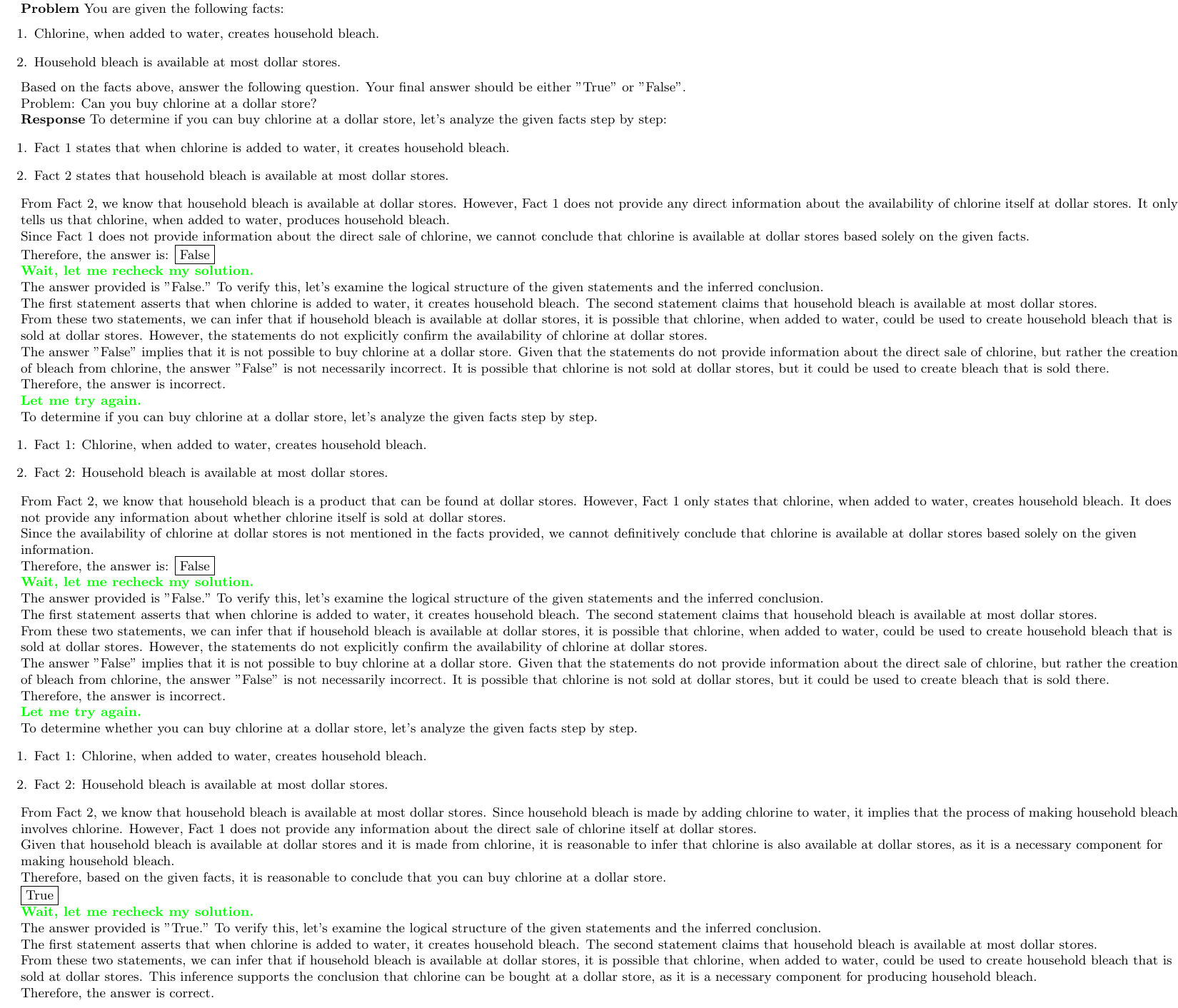}	
	\caption{StrategyQA Case.}
	\label{fig:sqa_case}
\end{figure*}

To intuitively demonstrate the effectiveness of our proposed method, we present the model's inference examples after RL on the MATH500 and StrategyQA datasets in the Figure \ref{fig:math_case} and Figure \ref{fig:sqa_case}.

\section{Other Discussion}
\subsection{Discussion on Potential Risk}
We have carefully considered potential risks associated with our work and found no significant concerns. Our approach, focused on enhancing LLM reasoning through self-verification and self-correction, does not introduce malicious or harmful effects, privacy issues, or security threats. Additionally, it does not contribute to biases, fairness concerns, or environmental impact. We believe our work is safe for responsible use in research.

\subsection{Use of AI Assistant}
In this work, we utilized an AI assistant solely for the purpose of refining and polishing the language of the manuscript. The AI assistant was employed to improve clarity, flow, and overall readability, ensuring the text adhered to academic writing standards. It was not involved in any data analysis, experimentation, or formulation of ideas. All research design, methodology, results, and conclusions were developed independently by the authors. The use of the AI assistant was limited to language enhancement and did not influence the content or scientific integrity of the work.
\end{document}